%% file: main.tex
\documentclass[10pt,twocolumn,letterpaper]{article}

\usepackage{cvpr}

\makeatletter
\@namedef{ver@everyshi.sty}{}
\makeatother

\usepackage{times}
\usepackage{epsfig}
\usepackage{graphicx}
\usepackage{amsmath}
\usepackage{amssymb}

\usepackage{epstopdf}
\usepackage{bm}
\usepackage{multirow}

\usepackage{pdfpages}
\usepackage{xspace}
\usepackage{times}
\usepackage{epsfig}
\usepackage{float}
\usepackage{xcolor}
\usepackage{tikz-qtree}
\usepackage{tikz}
\usepackage[draft]{todonotes}   
\usepackage{algorithm}
\usepackage{mathtools,xparse}
\usepackage{siunitx}
\usepackage{stmaryrd}
\usepackage{algpseudocode}
\usepackage{placeins}
\usepackage[export]{adjustbox}
\usepackage{diagbox}
\usepackage[acronym,nomain]{glossaries}

\usepackage{amsthm}
\usepackage{dirtree}
\usepackage{algorithmicx}
\usepackage{pgfplots}
\usepackage{enumitem}
\usepackage{tkz-fct} 
\usepackage{cellspace}

\usetikzlibrary{arrows.meta}
\usetikzlibrary{trees} 
\usetikzlibrary{shapes,arrows,decorations.markings}
\usetikzlibrary{spy}
\usetikzlibrary{calc}
\usetikzlibrary{intersections}
\usetikzlibrary{matrix}
\usetikzlibrary{datavisualization}

\usetikzlibrary{external}
\usepackage{caption}

\newcommand{\C}{\mathcal{C}}

\usepackage[pagebackref=true,breaklinks=true,letterpaper=true,colorlinks,bookmarks=false]{hyperref}

\makeatletter
\def\blfootnote{\xdef\@thefnmark{}\@footnotetext}
\makeatother

\cvprfinalcopy 

\def\cvprPaperID{4889} 

\ifcvprfinal\fi
\begin{document}

\title{Sub-frame Appearance and 6D Pose Estimation of Fast Moving Objects}

\author{Denys Rozumnyi$^{3,1}$
\and
Jan Kotera$^{2}$
\and
Filip \v{S}roubek$^{2}$
\and
Ji\v{r}\'{i} Matas$^{1}$ 
\and
{\normalsize $^{1}$Visual Recognition Group, Faculty of Electrical Engineering, Czech Technical University in Prague, Czech Republic }
\and
{\normalsize  $^{2}$UTIA, Czech Academy of Sciences, Prague, Czech Republic }
\and
{\normalsize $^{3}$Department of Computer Science, ETH Zurich }
}
%
\input{figures/teaser.tex}
\begin{abstract}


We propose a novel method  that tracks fast moving objects, mainly non-uniform spherical, in full 6 degrees of freedom, estimating simultaneously their 3D motion trajectory, 3D pose and object appearance changes with a time step that is a fraction of the video frame exposure time.
The sub-frame object localization and appearance estimation allows realistic temporal super-resolution and precise shape estimation.
The method, called  TbD-3D  (Tracking by Deblatting in 3D) relies on a novel reconstruction algorithm which solves a piece-wise deblurring and matting problem. The 3D rotation is estimated by minimizing the reprojection error.
As a second contribution, we present a new challenging dataset with fast moving objects that change their appearance and distance to the camera.
High speed camera recordings with zero lag between frame exposures were used to generate videos with different frame rates annotated with ground-truth trajectory and pose.

\end{abstract}

\input{figures/recon.tex}

\section{Introduction}

\input{sections/introduction.tex}

\section{Related Work}
\input{sections/related_work.tex}

\input{figures/recon2.tex}

\input{figures/traj3d.tex}

\section{Method}
\input{sections/method.tex}

\section{Experiments}
\input{sections/experiments.tex}

\section{Conclusion}
We proposed a method for estimating up to 6DoF trajectory of fast moving objects which are severely blurred by object motion. The assumption of a non-uniform spherical object is needed, otherwise only a 3D object location is estimated. The proposed TbD-3D method achieves good results on a newly created dataset of non-uniform FMOs with significant changes of appearance and distance to the camera within the sequence or even a frame. Sub-frame appearance estimation enables us to see deformations which last shorter than the exposure duration. We showed a more precise temporal super-resolution compared to the previous methods. The dataset and implementation will be made publicly available.

{\small
\bibliographystyle{ieee_fullname}
\bibliography{main}
}

\end{document}

%% file: figures/teaser.tex
\twocolumn[{%
\renewcommand\twocolumn[1][]{#1}%
\maketitle
\vspace*{-1cm}
\begin{center}
    \centering
	\includegraphics[width=0.95\textwidth]{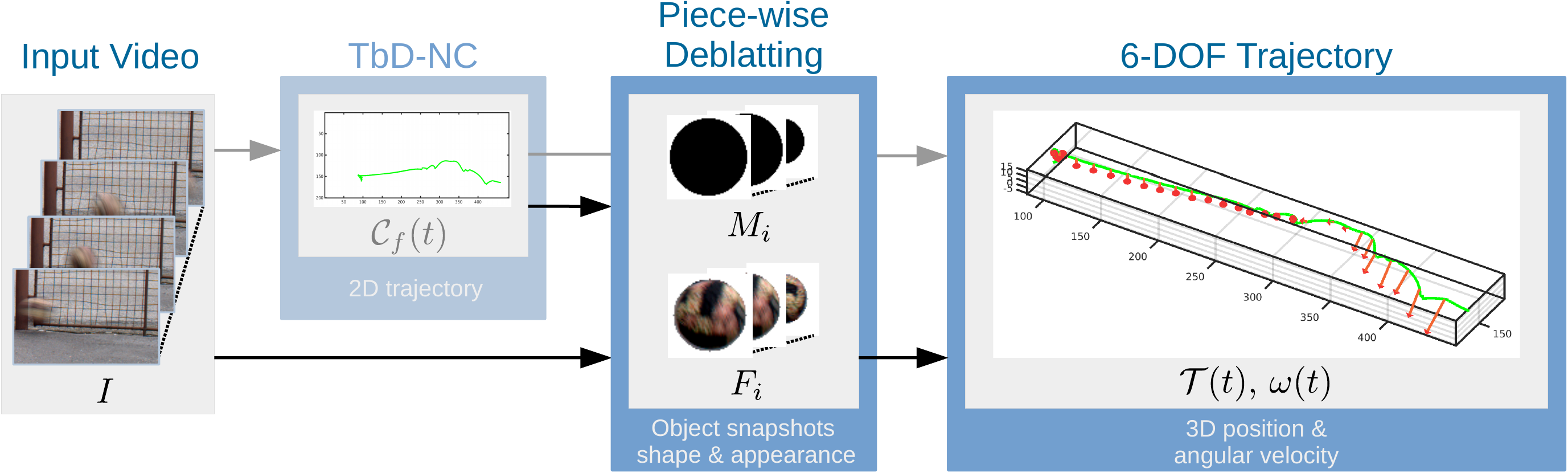}
 	\vspace*{-0.2cm}
    \captionof{figure}{Estimation of appearance, shape and 6D pose (3D position and rotation) of fast moving objects. The input video and 2D trajectories estimated by Non-Causal Tracking by Deblatting, TbD-NC \cite{tbdnc}, are processed by the proposed piece-wise deblatting that generates, with sub-frame temporal resolution, the object appearance and shape (snapshots), from which the complete 6-DOF trajectory is estimated.} 
\label{fig:teaser}
\end{center}
}]

%% file: figures/recon.tex
\begin{figure*}
\centering

\begin{tabular}{@{}c@{\hspace{0.5em}}c@{}c@{}c@{}c@{}c@{}c@{}c@{}c@{}}

\multirow{2}{*}[+1.25cm]{\includegraphics[height=0.18\textwidth]{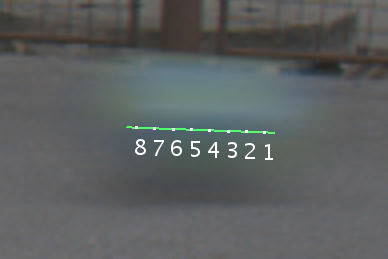}}  &

\includegraphics[width=0.09\textwidth]{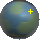}  & 
\includegraphics[width=0.09\textwidth]{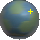}  & 
\includegraphics[width=0.09\textwidth]{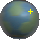}  & 
\includegraphics[width=0.09\textwidth]{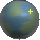}  & 
\includegraphics[width=0.09\textwidth]{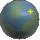}  & 
\includegraphics[width=0.09\textwidth]{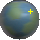}  & 
\includegraphics[width=0.09\textwidth]{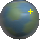}  & 
\includegraphics[width=0.09\textwidth]{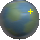}  \\

 &
\includegraphics[width=0.09\textwidth]{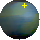}  &
\includegraphics[width=0.09\textwidth]{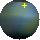}  &
\includegraphics[width=0.09\textwidth]{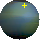}  &
\includegraphics[width=0.09\textwidth]{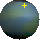}  &
\includegraphics[width=0.09\textwidth]{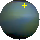}  &
\includegraphics[width=0.09\textwidth]{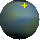}  &
\includegraphics[width=0.09\textwidth]{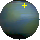}  &
\includegraphics[width=0.09\textwidth]{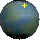}  \\

\multirow{2}{*}[+1.45cm]{\includegraphics[height=0.2\textwidth]{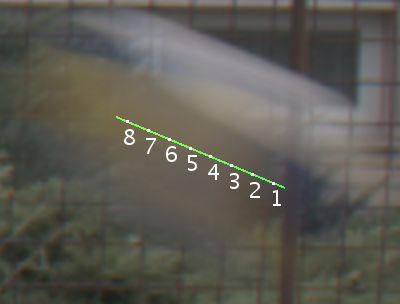}}  &

\includegraphics[width=0.09\textwidth]{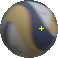}  & 
\includegraphics[width=0.09\textwidth]{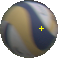}  & 
\includegraphics[width=0.09\textwidth]{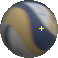}  & 
\includegraphics[width=0.09\textwidth]{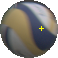}  & 
\includegraphics[width=0.09\textwidth]{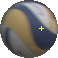}  & 
\includegraphics[width=0.09\textwidth]{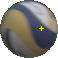}  & 
\includegraphics[width=0.09\textwidth]{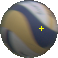}  & 
\includegraphics[width=0.09\textwidth]{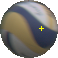}  \\

 &
\includegraphics[width=0.09\textwidth]{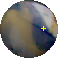}  &
\includegraphics[width=0.09\textwidth]{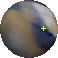}  &
\includegraphics[width=0.09\textwidth]{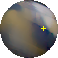}  &
\includegraphics[width=0.09\textwidth]{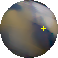}  &
\includegraphics[width=0.09\textwidth]{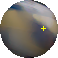}  &
\includegraphics[width=0.09\textwidth]{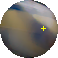}  &
\includegraphics[width=0.09\textwidth]{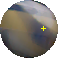}  &
\includegraphics[width=0.09\textwidth]{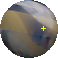}  \\

Input & 1 & 2 & 3 & 4 & 5 & 6 & 7 & 8 \\


\end{tabular}
\vspace*{-0.3cm}
\caption{Sub-frame appearance estimation of fast moving objects. Left: 30 fps input images with overlaid 2D projections of recovered 3D trajectories in green. White points correspond to time instants in the middle of high-speed camera frames. Right: cropped objects from a high-speed camera (top) and output of the proposed TbD-3D (bottom). 3D rotation is estimated by minimizing the reprojection error, assuming a spherical object. The estimated rotation axis is visualized by a yellow cross. 
}
\label{fig:recon} 
\end{figure*}

%% file: sections/introduction.tex
Visual tracking encompasses a broad class of problems that have received significant interest~\cite{vot2018,vot2019}. Current state-of-the-art methods employ a range of techniques, such as deep Siamese networks~\cite{Li_2018_CVPR,Valmadre_2017_CVPR} and discriminative correlation filters~\cite{xu2019learning,csrdcf}. The standard output of tracking methods is a 2D bounding box, either axis aligned or rotated.
Video segmentation methods output precise segmentation masks~\cite{6618931,8099545}. 

Recently, fast moving objects (FMOs) have been introduced as one of the problems in tracking~\cite{fmo}. Such objects are recorded as blurred streaks. They are common in sport videos and many other scenarios, such as videos of falling objects, hailstorm and flying insects, or more specialized ones, \eg visual navigation of microrobots in a magnetic field. 
To avoid FMOs and the related phenomena, one can use high-speed cameras operating at high frame rates,~\eg 240~fps or more. However, this brings additional requirements on resources, such as data transfer bandwidth and storage. When capturing such objects, camera settings have to be considered a priori before video acquisition.

The blurred trace of an object encodes information about its velocity, shape and appearance. Estimating these quantities should be thus in principle possible even from more affordable cameras with 30~fps, but it is a challenging task as the problem is heavily ill-posed. As shown in~\cite{fmo}, standard tracking methods do not perform well on FMOs.

For a fast moving object, a bounding box or a segmentation mask is not an adequate representation of its trajectory, as it travels a non-negligible path in a single frame. Such object may be localized more precisely, with a sub-frame accuracy.

Tracking by Deblatting (TbD)~\cite{tbd} was the first method to track fast moving objects by solving a joint \emph{debl}urring and m\emph{atting} (\emph{deblatting}) problem. These techniques are closer to blind deconvolution than to visual tracking methods. Non-causal post-processing proposed in~\cite{tbdnc} gives more precise and complete trajectories. The output of both above-mentioned methods is only a 2D trajectory.
They assume a 2D appearance and mask of an object to stay unchanged over the duration of a frame. 
This is equivalent to ignoring the 3D rotation of the object, the change of its distance to camera and of appearance due to the non-uniform light field, reflections, shadows or deformations.
Such simplifications are only adequate for objects with almost uniform texture and moving in a plane parallel to the camera projection plane. To date, the full nature of 3D object motion and appearance has not been considered, nor object location in 3D nor angular velocity in 3D. 

In this paper we are the first to estimate continuous-time sub-frame changes in appearance of the object. While solving for the shape and appearance, we recover the 3D rotation of the object and distance to the camera (currently we are able to handle only close to spherical object). The output of the proposed method is a continuous object pose with 6 degrees of freedom. 
The reconstruction pipeline is summarized in Figure~\ref{fig:teaser}.

We make the following \textbf{contributions}:

\begin{itemize}
\item We propose TbD-3D (Tracking by Deblatting in 3D)~--~the first method to reconstruct the appearance and the shape of blurred moving objects with sub-frame temporal resolution using piece-wise deblatting. We call these reconstructions snapshots (as in Figure~\ref{fig:recon}).
\item The method estimates continuous-time pose with 6 degrees of freedom (3D location and rotation) for non-uniform spherical objects. The rotation is estimated by a new method which minimizes the reprojection error.
\item We collect and make available a new challenging dataset with fast moving objects that change their appearance and distance to the camera. High speed camera with zero lag between frame exposures is used to generate videos with different low frame rates annotated with ground-truth trajectory and pose data.
\item Sub-frame reconstruction accuracy on object deformations that occur during contact with other objects is demonstrated.
\end{itemize}

%% file: sections/related_work.tex
Detection and tracking of fast moving objects was introduced by Rozumnyi~\etal~\cite{fmo}. Their work was limited by several assumptions on object trajectory and appearance, such as linear trajectory parallel to the camera projection plane, uniform 2D appearance of the object, high contrast to the background and no contact of the moving object with other objects. Some of these assumptions were relaxed in a method called Tracking by Deblatting (TbD)~\cite{tbd}, which tracks fast moving objects by solving a deblurring problem in every frame and processing the video in a causal manner. TbD outperforms the previous approach by a wide margin, yet trajectories estimated at adjacent frames are independent and the final trajectory for the whole sequence is a set of segments. These limitations were addressed by a follow-up method called non-causal tracking by deblatting (TbD-NC)~\cite{tbdnc}. TbD-NC takes the output of TbD and estimates the final trajectory which is continuous over the duration of the whole sequence. 

All these methods assume that the object trajectory is parallel to the camera plane and that the object appearance is static within one frame (no rotation). The appearance can change between frames, but in arbitrary fashion 
as a long-term appearance template was learned online. The only exception is the work of Kotera and \v{S}roubek~\cite{kotera_fmo} that estimates object rotation, yet only 2D in-plane rotation is considered and the object shape is assumed to be known. The method is thus applicable only to nearly flat objects moving on a flat surface.

Deep learning has been applied to motion deblurring of videos \cite{Wieschollek2017,Su2017} and to the generation of intermediate short-exposure frames \cite{Jin2018}. Their proposed convolutional neural networks are trained only on small blurs; blur parameters are not available as they are not directly estimated. 
Tracking methods that consider blurred objects in \cite{Park2009, Xu2016_tracking} assume object motion that is approximately linear and relatively small compared to the object size. Alpha blending of the tracked object with the background is ignored and their output per frame is only a bounding box, which is insufficient for fast moving objects.


The tracking methods \cite{fmo,tbd,tbdnc} for fast moving objects use an image formation model that is defined as
\begin{equation}
	\label{eq:acquisition_model}
	I = H*F + (1-H*M)B\,
\end{equation}
for a single color video frame $I$. The formation model is a mixture of two components. The first component is the object appearance $F$ (after projection to the image plane) blurred by motion along the object trajectory, which is represented as a blur kernel $H$. The second part is the background $B$ attenuated by object occlusion, where $M$, equivalent to the indicator function of $F$, is the object shape after projection to the image plane. Following~\cite{tbd}, the blur is simplified to a 2D convolution. The background $B$ is estimated as a median of the last 5 frames. They assume either an almost static camera or a stabilized sequence.

The output of TbD-NC~\cite{tbdnc} is a 2D object trajectory $\C_f(t)$: $[0,N]\to\mathbb{R}^2$ in an analytical form where $N$ is the number of frames in the video sequence. This output is then used as an input to the proposed TbD-3D method.

%% file: figures/recon2.tex
\begin{figure}
\centering

\begin{tabular}{@{}cc@{}c@{}c@{}c@{}}

\includegraphics[height=0.085\textwidth]{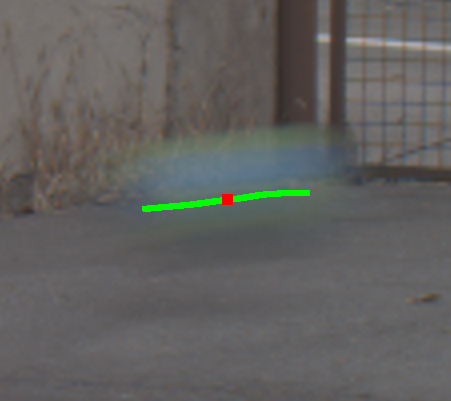} &  
\includegraphics[height=0.085\textwidth]{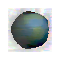} &  
\includegraphics[height=0.085\textwidth]{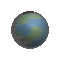} &  
\includegraphics[height=0.085\textwidth]{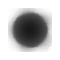} &  
\includegraphics[height=0.085\textwidth]{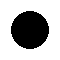} \\

\includegraphics[height=0.085\textwidth]{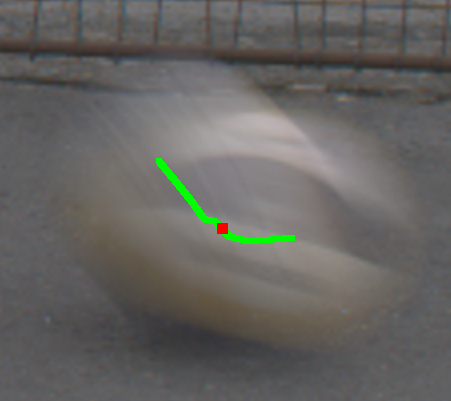} &  
\includegraphics[height=0.085\textwidth]{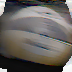} &  
\includegraphics[height=0.085\textwidth]{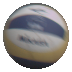} &  
\includegraphics[height=0.085\textwidth]{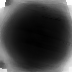} &  
\includegraphics[height=0.085\textwidth]{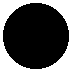} \\

\includegraphics[height=0.085\textwidth]{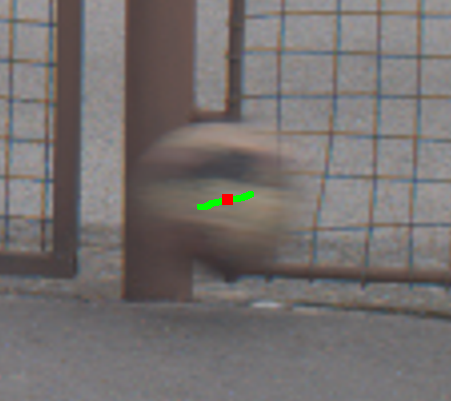} &  
\includegraphics[height=0.085\textwidth]{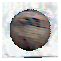} &  
\includegraphics[height=0.085\textwidth]{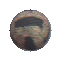} &  
\includegraphics[height=0.085\textwidth]{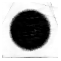} &  
\includegraphics[height=0.085\textwidth]{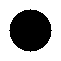} \\

\includegraphics[height=0.085\textwidth]{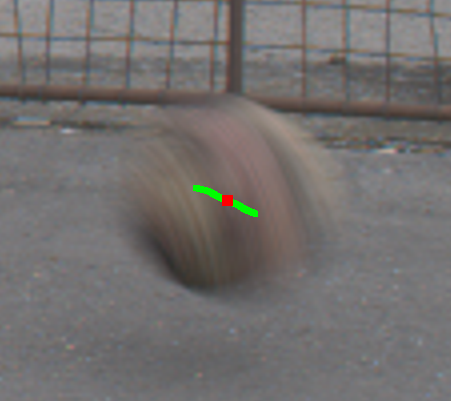} &  
\includegraphics[height=0.085\textwidth]{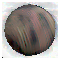} &  
\includegraphics[height=0.085\textwidth]{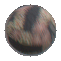} &  
\includegraphics[height=0.085\textwidth]{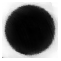} &  
\includegraphics[height=0.085\textwidth]{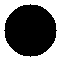} \\

Input & $F$ & $F^*$ & $M$ & $M^*$ \\

\end{tabular}
\vspace*{-0.3cm}
\caption{TbD-3D applied to 30~fps videos compared to images from a high-speed camera at 240~fps (marked with $*$). $F$: snapshots of object appearance estimates of fast moving objects. Each row corresponds to one sub-frame instant (red dot on a green trajectory) of the input frame on the left. For visualization purposes, the masks $M$ are inverted.}

\label{fig:recon2} 
\end{figure}

%% file: figures/traj3d.tex
\begin{figure*}
\centering
\begin{tabular}{@{}c@{}c@{}c@{}}
\input{teximgs/depth2.tex}    & 
\input{teximgs/depth2_o.tex}& 
\input{teximgs/depth2_e.tex} \\

\input{teximgs/depthf1.tex} &
\input{teximgs/depthf1_o.tex} &
\input{teximgs/depthf1_e.tex} \\

\input{teximgs/depthb2.tex} & 
\input{teximgs/depthb2_o.tex} &
\input{teximgs/depthb2_e.tex}\\

Ground Truth & TbD-3D with Oracle & TbD-3D \\

\end{tabular}
\vspace*{-0.3cm}
\caption{3D trajectory estimation for selected sequences from the TbD-3D dataset. Curve thickness codes distance from the object to the camera (thicker curve means that the object is closer to the camera). TbD-3D with Oracle means that the 2D trajectory is estimated from the original high-speed footage and only the third dimension is estimated. Otherwise, the output of TbD-NC~\cite{tbdnc} is used as the 2D trajectory. Sequences correspond to 30~fps. }
\label{fig:tbd3d_imgs}
\end{figure*}

%% file: teximgs/depth2_e.tex
\resizebox {0.333\textwidth}{!} {\begin{tikzpicture} 
\begin{axis}[y dir=reverse, 
 xmin=1,xmax=960, 
 ymin=1,ymax=400, 
 xticklabels = \empty, yticklabels = \empty, 
 grid=none, axis equal image] 
\addplot graphics[xmin=1,xmax=960,ymin=1,ymax=400] {imgs/thumbnails/tbd/depth2.png}; 
\addplot [>={Latex[length=1.5mm,width=0.5mm,angle'=25,open,round]},,domain=1:1.2,samples=2,line width=2.2108,color=blue]({765.5004 + -37.7091*x + 1.5598*x^2},{329.0395 + -30.1187*x + 1.4694*x^2});  
\addplot [>={Latex[length=1.5mm,width=0.5mm,angle'=25,open,round]},,domain=1.2:1.4,samples=2,line width=2.1293,color=blue]({765.5004 + -37.7091*x + 1.5598*x^2},{329.0395 + -30.1187*x + 1.4694*x^2});  
\addplot [>={Latex[length=1.5mm,width=0.5mm,angle'=25,open,round]},,domain=1.4:1.6,samples=3,line width=2.042,color=blue]({765.5004 + -37.7091*x + 1.5598*x^2},{329.0395 + -30.1187*x + 1.4694*x^2});  
\addplot [>={Latex[length=1.5mm,width=0.5mm,angle'=25,open,round]},,domain=1.6:1.8,samples=2,line width=1.952,color=blue]({765.5004 + -37.7091*x + 1.5598*x^2},{329.0395 + -30.1187*x + 1.4694*x^2});  
\addplot [>={Latex[length=1.5mm,width=0.5mm,angle'=25,open,round]},,domain=1.8:2,samples=2,line width=1.8453,color=blue]({765.5004 + -37.7091*x + 1.5598*x^2},{329.0395 + -30.1187*x + 1.4694*x^2});  
\addplot [>={Latex[length=1.5mm,width=0.5mm,angle'=25,open,round]},,domain=2:2.2,samples=3,line width=1.737,color=blue]({765.5004 + -37.7091*x + 1.5598*x^2},{329.0395 + -30.1187*x + 1.4694*x^2});  
\addplot [>={Latex[length=1.5mm,width=0.5mm,angle'=25,open,round]},,domain=2.2:2.4,samples=3,line width=1.634,color=blue]({765.5004 + -37.7091*x + 1.5598*x^2},{329.0395 + -30.1187*x + 1.4694*x^2});  
\addplot [>={Latex[length=1.5mm,width=0.5mm,angle'=25,open,round]},,domain=2.4:2.6,samples=2,line width=1.5752,color=blue]({765.5004 + -37.7091*x + 1.5598*x^2},{329.0395 + -30.1187*x + 1.4694*x^2});  
\addplot [>={Latex[length=1.5mm,width=0.5mm,angle'=25,open,round]},,domain=2.6:2.8,samples=2,line width=1.5309,color=blue]({765.5004 + -37.7091*x + 1.5598*x^2},{329.0395 + -30.1187*x + 1.4694*x^2});  
\addplot [>={Latex[length=1.5mm,width=0.5mm,angle'=25,open,round]},,domain=2.8:3,samples=3,line width=1.4859,color=blue]({765.5004 + -37.7091*x + 1.5598*x^2},{329.0395 + -30.1187*x + 1.4694*x^2});  
\addplot [>={Latex[length=1.5mm,width=0.5mm,angle'=25,open,round]},,domain=3:3.2,samples=3,line width=1.4937,color=blue]({765.5004 + -37.7091*x + 1.5598*x^2},{329.0395 + -30.1187*x + 1.4694*x^2});  
\addplot [>={Latex[length=1.5mm,width=0.5mm,angle'=25,open,round]},,domain=3.2:3.4,samples=3,line width=1.4918,color=blue]({765.5004 + -37.7091*x + 1.5598*x^2},{329.0395 + -30.1187*x + 1.4694*x^2});  
\addplot [>={Latex[length=1.5mm,width=0.5mm,angle'=25,open,round]},,domain=3.4:3.6,samples=2,line width=1.4428,color=blue]({765.5004 + -37.7091*x + 1.5598*x^2},{329.0395 + -30.1187*x + 1.4694*x^2});  
\addplot [>={Latex[length=1.5mm,width=0.5mm,angle'=25,open,round]},,domain=3.6:3.8,samples=3,line width=1.3757,color=blue]({765.5004 + -37.7091*x + 1.5598*x^2},{329.0395 + -30.1187*x + 1.4694*x^2});  
\addplot [>={Latex[length=1.5mm,width=0.5mm,angle'=25,open,round]},,domain=3.8:4,samples=2,line width=1.2767,color=blue]({765.5004 + -37.7091*x + 1.5598*x^2},{329.0395 + -30.1187*x + 1.4694*x^2});  
\addplot [>={Latex[length=1.5mm,width=0.5mm,angle'=25,open,round]},,domain=4:4.2,samples=2,line width=1.1895,color=blue]({765.5004 + -37.7091*x + 1.5598*x^2},{329.0395 + -30.1187*x + 1.4694*x^2});  
\addplot [>={Latex[length=1.5mm,width=0.5mm,angle'=25,open,round]},,domain=4.2:4.4,samples=3,line width=1.1277,color=blue]({765.5004 + -37.7091*x + 1.5598*x^2},{329.0395 + -30.1187*x + 1.4694*x^2});  
\addplot [>={Latex[length=1.5mm,width=0.5mm,angle'=25,open,round]},,domain=4.4:4.6,samples=2,line width=1.0643,color=blue]({765.5004 + -37.7091*x + 1.5598*x^2},{329.0395 + -30.1187*x + 1.4694*x^2});  
\addplot [>={Latex[length=1.5mm,width=0.5mm,angle'=25,open,round]},,domain=4.6:4.8,samples=3,line width=1.0077,color=blue]({765.5004 + -37.7091*x + 1.5598*x^2},{329.0395 + -30.1187*x + 1.4694*x^2});  
\addplot [>={Latex[length=1.5mm,width=0.5mm,angle'=25,open,round]},,domain=4.8:5,samples=3,line width=0.95308,color=blue]({765.5004 + -37.7091*x + 1.5598*x^2},{329.0395 + -30.1187*x + 1.4694*x^2});  
\addplot [>={Latex[length=1.5mm,width=0.5mm,angle'=25,open,round]},,domain=5:5.2,samples=3,line width=0.90056,color=blue]({765.5004 + -37.7091*x + 1.5598*x^2},{329.0395 + -30.1187*x + 1.4694*x^2});  
\addplot [>={Latex[length=1.5mm,width=0.5mm,angle'=25,open,round]},,domain=5.2:5.4,samples=3,line width=0.85407,color=blue]({765.5004 + -37.7091*x + 1.5598*x^2},{329.0395 + -30.1187*x + 1.4694*x^2});  
\addplot [>={Latex[length=1.5mm,width=0.5mm,angle'=25,open,round]},,domain=5.4:5.6,samples=2,line width=0.82147,color=blue]({765.5004 + -37.7091*x + 1.5598*x^2},{329.0395 + -30.1187*x + 1.4694*x^2});  
\addplot [>={Latex[length=1.5mm,width=0.5mm,angle'=25,open,round]},,domain=5.6:5.8,samples=3,line width=0.789,color=blue]({765.5004 + -37.7091*x + 1.5598*x^2},{329.0395 + -30.1187*x + 1.4694*x^2});  
\addplot [>={Latex[length=1.5mm,width=0.5mm,angle'=25,open,round]},,domain=5.8:6,samples=3,line width=0.75315,color=blue]({765.5004 + -37.7091*x + 1.5598*x^2},{329.0395 + -30.1187*x + 1.4694*x^2});  
\addplot [>={Latex[length=1.5mm,width=0.5mm,angle'=25,open,round]},,domain=6:6.2,samples=3,line width=0.72161,color=blue]({765.5004 + -37.7091*x + 1.5598*x^2},{329.0395 + -30.1187*x + 1.4694*x^2});  
\addplot [>={Latex[length=1.5mm,width=0.5mm,angle'=25,open,round]},,domain=6.2:6.4,samples=3,line width=0.69738,color=blue]({765.5004 + -37.7091*x + 1.5598*x^2},{329.0395 + -30.1187*x + 1.4694*x^2});  
\addplot [>={Latex[length=1.5mm,width=0.5mm,angle'=25,open,round]},,domain=6.4:6.6,samples=2,line width=0.6787,color=blue]({765.5004 + -37.7091*x + 1.5598*x^2},{329.0395 + -30.1187*x + 1.4694*x^2});  
\addplot [>={Latex[length=1.5mm,width=0.5mm,angle'=25,open,round]},,domain=6.6:6.8,samples=3,line width=0.65926,color=blue]({765.5004 + -37.7091*x + 1.5598*x^2},{329.0395 + -30.1187*x + 1.4694*x^2});  
\addplot [>={Latex[length=1.5mm,width=0.5mm,angle'=25,open,round]},,domain=6.8:7,samples=3,line width=0.63887,color=blue]({765.5004 + -37.7091*x + 1.5598*x^2},{329.0395 + -30.1187*x + 1.4694*x^2});  
\addplot [>={Latex[length=1.5mm,width=0.5mm,angle'=25,open,round]},,domain=0:1,samples=2,line width=0.63887,color=blue]({577.9651 + 1.3642e-12*x},{190.2098 + 1.9327e-12*x});  
\addplot [>={Latex[length=1.5mm,width=0.5mm,angle'=25,open,round]},,domain=0:1,samples=2,line width=0.63887,color=blue]({577.9651 + -13.9651*x},{190.2098 + 4.7902*x});  
\addplot [>={Latex[length=1.5mm,width=0.5mm,angle'=25,open,round]},,domain=0:1,samples=2,line width=0.63887,color=blue]({564 + -16.8014*x},{195 + -9.4771*x});  
\addplot [>={Latex[length=1.5mm,width=0.5mm,angle'=25,open,round]},,domain=0:1,samples=2,line width=0.63887,color=blue]({547.1986 + -60.4014*x},{185.5229 + -13.6218*x});  
\addplot [>={Latex[length=1.5mm,width=0.5mm,angle'=25,open,round]},,domain=8:8.2,samples=2,line width=0.56712,color=blue]({18.8622 + 98.5632*x + -5.0089*x^2},{6.9318 + 36.2167*x + -1.9494*x^2});  
\addplot [>={Latex[length=1.5mm,width=0.5mm,angle'=25,open,round]},,domain=8.2:8.4,samples=3,line width=0.55434,color=blue]({18.8622 + 98.5632*x + -5.0089*x^2},{6.9318 + 36.2167*x + -1.9494*x^2});  
\addplot [>={Latex[length=1.5mm,width=0.5mm,angle'=25,open,round]},,domain=8.4:8.6,samples=2,line width=0.53828,color=blue]({18.8622 + 98.5632*x + -5.0089*x^2},{6.9318 + 36.2167*x + -1.9494*x^2});  
\addplot [>={Latex[length=1.5mm,width=0.5mm,angle'=25,open,round]},,domain=8.6:8.8,samples=3,line width=0.51994,color=blue]({18.8622 + 98.5632*x + -5.0089*x^2},{6.9318 + 36.2167*x + -1.9494*x^2});  
\addplot [>={Latex[length=1.5mm,width=0.5mm,angle'=25,open,round]},,domain=8.8:9,samples=2,line width=0.50203,color=blue]({18.8622 + 98.5632*x + -5.0089*x^2},{6.9318 + 36.2167*x + -1.9494*x^2});  
\addplot [>={Latex[length=1.5mm,width=0.5mm,angle'=25,open,round]},,domain=9:9.2,samples=2,line width=0.49129,color=blue]({18.8622 + 98.5632*x + -5.0089*x^2},{6.9318 + 36.2167*x + -1.9494*x^2});  
\addplot [>={Latex[length=1.5mm,width=0.5mm,angle'=25,open,round]},,domain=9.2:9.4,samples=3,line width=0.48757,color=blue]({18.8622 + 98.5632*x + -5.0089*x^2},{6.9318 + 36.2167*x + -1.9494*x^2});  
\addplot [>={Latex[length=1.5mm,width=0.5mm,angle'=25,open,round]},,domain=9.4:9.6,samples=2,line width=0.48385,color=blue]({18.8622 + 98.5632*x + -5.0089*x^2},{6.9318 + 36.2167*x + -1.9494*x^2});  
\addplot [>={Latex[length=1.5mm,width=0.5mm,angle'=25,open,round]},,domain=9.6:9.8,samples=3,line width=0.48014,color=blue]({18.8622 + 98.5632*x + -5.0089*x^2},{6.9318 + 36.2167*x + -1.9494*x^2});  
\addplot [>={Latex[length=1.5mm,width=0.5mm,angle'=25,open,round]},,domain=9.8:10,samples=2,line width=0.47642,color=blue]({18.8622 + 98.5632*x + -5.0089*x^2},{6.9318 + 36.2167*x + -1.9494*x^2});  
\addplot [>={Latex[length=1.5mm,width=0.5mm,angle'=25,open,round]},,domain=10:10.2,samples=2,line width=0.47323,color=blue]({18.8622 + 98.5632*x + -5.0089*x^2},{6.9318 + 36.2167*x + -1.9494*x^2});  
\addplot [>={Latex[length=1.5mm,width=0.5mm,angle'=25,open,round]},,domain=10.2:10.4,samples=3,line width=0.47056,color=blue]({18.8622 + 98.5632*x + -5.0089*x^2},{6.9318 + 36.2167*x + -1.9494*x^2});  
\addplot [>={Latex[length=1.5mm,width=0.5mm,angle'=25,open,round]},,domain=10.4:10.6,samples=2,line width=0.46789,color=blue]({18.8622 + 98.5632*x + -5.0089*x^2},{6.9318 + 36.2167*x + -1.9494*x^2});  
\addplot [>={Latex[length=1.5mm,width=0.5mm,angle'=25,open,round]},,domain=10.6:10.8,samples=3,line width=0.46522,color=blue]({18.8622 + 98.5632*x + -5.0089*x^2},{6.9318 + 36.2167*x + -1.9494*x^2});  
\addplot [>={Latex[length=1.5mm,width=0.5mm,angle'=25,open,round]},,domain=10.8:11,samples=2,line width=0.46255,color=blue]({18.8622 + 98.5632*x + -5.0089*x^2},{6.9318 + 36.2167*x + -1.9494*x^2});  
\addplot [>={Latex[length=1.5mm,width=0.5mm,angle'=25,open,round]},,domain=0:1,samples=2,line width=0.46255,color=blue]({496.9786 + -2.2737e-13*x},{169.433 + -5.6843e-14*x});  
\addplot [>={Latex[length=1.5mm,width=0.5mm,angle'=25,open,round]},,domain=0:1,samples=2,line width=0.46255,color=blue]({496.9786 + -17.9786*x},{169.433 + -16.433*x});  
\addplot [>={Latex[length=1.5mm,width=0.5mm,angle'=25,open,round]},,domain=0:1,samples=2,line width=0.46255,color=blue]({479 + -12.9627*x},{153 + 2.2216*x});  
\addplot [>={Latex[length=1.5mm,width=0.5mm,angle'=25,open,round]},,domain=12:12.2,samples=2,line width=0.56715,color=blue]({550.4182 + -7.0317*x},{159.2597 + -0.33651*x});  
\addplot [>={Latex[length=1.5mm,width=0.5mm,angle'=25,open,round]},,domain=12.2:12.4,samples=3,line width=0.57178,color=blue]({550.4182 + -7.0317*x},{159.2597 + -0.33651*x});  
\addplot [>={Latex[length=1.5mm,width=0.5mm,angle'=25,open,round]},,domain=12.4:12.6,samples=2,line width=0.5764,color=blue]({550.4182 + -7.0317*x},{159.2597 + -0.33651*x});  
\addplot [>={Latex[length=1.5mm,width=0.5mm,angle'=25,open,round]},,domain=12.6:12.8,samples=3,line width=0.58103,color=blue]({550.4182 + -7.0317*x},{159.2597 + -0.33651*x});  
\addplot [>={Latex[length=1.5mm,width=0.5mm,angle'=25,open,round]},,domain=12.8:13,samples=2,line width=0.58565,color=blue]({550.4182 + -7.0317*x},{159.2597 + -0.33651*x});  
\addplot [>={Latex[length=1.5mm,width=0.5mm,angle'=25,open,round]},,domain=0:1,samples=2,line width=0.58565,color=blue]({459.0055 + -5.6843e-14*x},{154.8851 + -2.8422e-14*x});  
\addplot [>={Latex[length=1.5mm,width=0.5mm,angle'=25,open,round]},,domain=0:1,samples=2,line width=0.58565,color=blue]({459.0055 + -106.0055*x},{154.8851 + 3.1149*x});  
\addplot [>={Latex[length=1.5mm,width=0.5mm,angle'=25,open,round]},,domain=0:1,samples=2,line width=0.58565,color=blue]({353 + -0.91711*x},{158 + -0.38307*x});  
\addplot [>={Latex[length=1.5mm,width=0.5mm,angle'=25,open,round]},,domain=0:1,samples=2,line width=0.58565,color=blue]({352.0829 + 2.2169e-12*x},{157.6169 + 8.1855e-12*x});  
\addplot [>={Latex[length=1.5mm,width=0.5mm,angle'=25,open,round]},,domain=18:18.2,samples=2,line width=1.0875,color=blue]({129.7819 + 22.0355*x + -0.53808*x^2},{464.5663 + -34.8476*x + 0.9886*x^2});  
\addplot [>={Latex[length=1.5mm,width=0.5mm,angle'=25,open,round]},,domain=18.2:18.4,samples=2,line width=1.1035,color=blue]({129.7819 + 22.0355*x + -0.53808*x^2},{464.5663 + -34.8476*x + 0.9886*x^2});  
\addplot [>={Latex[length=1.5mm,width=0.5mm,angle'=25,open,round]},,domain=18.4:18.6,samples=3,line width=1.1196,color=blue]({129.7819 + 22.0355*x + -0.53808*x^2},{464.5663 + -34.8476*x + 0.9886*x^2});  
\addplot [>={Latex[length=1.5mm,width=0.5mm,angle'=25,open,round]},,domain=18.6:18.8,samples=2,line width=1.1357,color=blue]({129.7819 + 22.0355*x + -0.53808*x^2},{464.5663 + -34.8476*x + 0.9886*x^2});  
\addplot [>={Latex[length=1.5mm,width=0.5mm,angle'=25,open,round]},,domain=18.8:19,samples=2,line width=1.1517,color=blue]({129.7819 + 22.0355*x + -0.53808*x^2},{464.5663 + -34.8476*x + 0.9886*x^2});  
\addplot [>={Latex[length=1.5mm,width=0.5mm,angle'=25,open,round]},,domain=19:19.2,samples=2,line width=1.1645,color=blue]({129.7819 + 22.0355*x + -0.53808*x^2},{464.5663 + -34.8476*x + 0.9886*x^2});  
\addplot [>={Latex[length=1.5mm,width=0.5mm,angle'=25,open,round]},,domain=19.2:19.4,samples=2,line width=1.1741,color=blue]({129.7819 + 22.0355*x + -0.53808*x^2},{464.5663 + -34.8476*x + 0.9886*x^2});  
\addplot [>={Latex[length=1.5mm,width=0.5mm,angle'=25,open,round]},,domain=19.4:19.6,samples=3,line width=1.1836,color=blue]({129.7819 + 22.0355*x + -0.53808*x^2},{464.5663 + -34.8476*x + 0.9886*x^2});  
\addplot [>={Latex[length=1.5mm,width=0.5mm,angle'=25,open,round]},,domain=19.6:19.8,samples=2,line width=1.1932,color=blue]({129.7819 + 22.0355*x + -0.53808*x^2},{464.5663 + -34.8476*x + 0.9886*x^2});  
\addplot [>={Latex[length=1.5mm,width=0.5mm,angle'=25,open,round]},,domain=19.8:20,samples=2,line width=1.2027,color=blue]({129.7819 + 22.0355*x + -0.53808*x^2},{464.5663 + -34.8476*x + 0.9886*x^2});  
\addplot [>={Latex[length=1.5mm,width=0.5mm,angle'=25,open,round]},,domain=20:20.2,samples=2,line width=1.2123,color=blue]({129.7819 + 22.0355*x + -0.53808*x^2},{464.5663 + -34.8476*x + 0.9886*x^2});  
\addplot [>={Latex[length=1.5mm,width=0.5mm,angle'=25,open,round]},,domain=20.2:20.4,samples=2,line width=1.2218,color=blue]({129.7819 + 22.0355*x + -0.53808*x^2},{464.5663 + -34.8476*x + 0.9886*x^2});  
\addplot [>={Latex[length=1.5mm,width=0.5mm,angle'=25,open,round]},,domain=20.4:20.6,samples=3,line width=1.2314,color=blue]({129.7819 + 22.0355*x + -0.53808*x^2},{464.5663 + -34.8476*x + 0.9886*x^2});  
\addplot [>={Latex[length=1.5mm,width=0.5mm,angle'=25,open,round]},,domain=20.6:20.8,samples=2,line width=1.241,color=blue]({129.7819 + 22.0355*x + -0.53808*x^2},{464.5663 + -34.8476*x + 0.9886*x^2});  
\addplot [>={Latex[length=1.5mm,width=0.5mm,angle'=25,open,round]},,domain=20.8:21,samples=2,line width=1.2506,color=blue]({129.7819 + 22.0355*x + -0.53808*x^2},{464.5663 + -34.8476*x + 0.9886*x^2});  
\addplot [>={Latex[length=1.5mm,width=0.5mm,angle'=25,open,round]},,domain=21:21.2,samples=2,line width=1.2599,color=blue]({129.7819 + 22.0355*x + -0.53808*x^2},{464.5663 + -34.8476*x + 0.9886*x^2});  
\addplot [>={Latex[length=1.5mm,width=0.5mm,angle'=25,open,round]},,domain=21.2:21.4,samples=2,line width=1.2691,color=blue]({129.7819 + 22.0355*x + -0.53808*x^2},{464.5663 + -34.8476*x + 0.9886*x^2});  
\addplot [>={Latex[length=1.5mm,width=0.5mm,angle'=25,open,round]},,domain=21.4:21.6,samples=3,line width=1.2782,color=blue]({129.7819 + 22.0355*x + -0.53808*x^2},{464.5663 + -34.8476*x + 0.9886*x^2});  
\addplot [>={Latex[length=1.5mm,width=0.5mm,angle'=25,open,round]},,domain=21.6:21.8,samples=2,line width=1.2874,color=blue]({129.7819 + 22.0355*x + -0.53808*x^2},{464.5663 + -34.8476*x + 0.9886*x^2});  
\addplot [>={Latex[length=1.5mm,width=0.5mm,angle'=25,open,round]},,domain=21.8:22,samples=2,line width=1.2965,color=blue]({129.7819 + 22.0355*x + -0.53808*x^2},{464.5663 + -34.8476*x + 0.9886*x^2});  
\addplot [>={Latex[length=1.5mm,width=0.5mm,angle'=25,open,round]},,domain=22:22.2,samples=2,line width=1.3139,color=blue]({129.7819 + 22.0355*x + -0.53808*x^2},{464.5663 + -34.8476*x + 0.9886*x^2});  
\addplot [>={Latex[length=1.5mm,width=0.5mm,angle'=25,open,round]},,domain=22.2:22.4,samples=2,line width=1.3394,color=blue]({129.7819 + 22.0355*x + -0.53808*x^2},{464.5663 + -34.8476*x + 0.9886*x^2});  
\addplot [>={Latex[length=1.5mm,width=0.5mm,angle'=25,open,round]},,domain=22.4:22.6,samples=3,line width=1.365,color=blue]({129.7819 + 22.0355*x + -0.53808*x^2},{464.5663 + -34.8476*x + 0.9886*x^2});  
\addplot [>={Latex[length=1.5mm,width=0.5mm,angle'=25,open,round]},,domain=22.6:22.8,samples=2,line width=1.3602,color=blue]({129.7819 + 22.0355*x + -0.53808*x^2},{464.5663 + -34.8476*x + 0.9886*x^2});  
\addplot [>={Latex[length=1.5mm,width=0.5mm,angle'=25,open,round]},,domain=22.8:23,samples=2,line width=1.3555,color=blue]({129.7819 + 22.0355*x + -0.53808*x^2},{464.5663 + -34.8476*x + 0.9886*x^2});  
\addplot [>={Latex[length=1.5mm,width=0.5mm,angle'=25,open,round]},,domain=23:23.2,samples=2,line width=1.3509,color=blue]({129.7819 + 22.0355*x + -0.53808*x^2},{464.5663 + -34.8476*x + 0.9886*x^2});  
\addplot [>={Latex[length=1.5mm,width=0.5mm,angle'=25,open,round]},,domain=23.2:23.4,samples=2,line width=1.3465,color=blue]({129.7819 + 22.0355*x + -0.53808*x^2},{464.5663 + -34.8476*x + 0.9886*x^2});  
\addplot [>={Latex[length=1.5mm,width=0.5mm,angle'=25,open,round]},,domain=23.4:23.6,samples=3,line width=1.3422,color=blue]({129.7819 + 22.0355*x + -0.53808*x^2},{464.5663 + -34.8476*x + 0.9886*x^2});  
\addplot [>={Latex[length=1.5mm,width=0.5mm,angle'=25,open,round]},,domain=23.6:23.8,samples=2,line width=1.3415,color=blue]({129.7819 + 22.0355*x + -0.53808*x^2},{464.5663 + -34.8476*x + 0.9886*x^2});  
\addplot [>={Latex[length=1.5mm,width=0.5mm,angle'=25,open,round]},,domain=23.8:24,samples=2,line width=1.3408,color=blue]({129.7819 + 22.0355*x + -0.53808*x^2},{464.5663 + -34.8476*x + 0.9886*x^2});  
\addplot [>={Latex[length=1.5mm,width=0.5mm,angle'=25,open,round]},,domain=24:24.2,samples=2,line width=1.3456,color=blue]({129.7819 + 22.0355*x + -0.53808*x^2},{464.5663 + -34.8476*x + 0.9886*x^2});  
\addplot [>={Latex[length=1.5mm,width=0.5mm,angle'=25,open,round]},,domain=24.2:24.4,samples=2,line width=1.3571,color=blue]({129.7819 + 22.0355*x + -0.53808*x^2},{464.5663 + -34.8476*x + 0.9886*x^2});  
\addplot [>={Latex[length=1.5mm,width=0.5mm,angle'=25,open,round]},,domain=24.4:24.6,samples=3,line width=1.3722,color=blue]({129.7819 + 22.0355*x + -0.53808*x^2},{464.5663 + -34.8476*x + 0.9886*x^2});  
\addplot [>={Latex[length=1.5mm,width=0.5mm,angle'=25,open,round]},,domain=24.6:24.8,samples=2,line width=1.3894,color=blue]({129.7819 + 22.0355*x + -0.53808*x^2},{464.5663 + -34.8476*x + 0.9886*x^2});  
\addplot [>={Latex[length=1.5mm,width=0.5mm,angle'=25,open,round]},,domain=24.8:25,samples=2,line width=1.403,color=blue]({129.7819 + 22.0355*x + -0.53808*x^2},{464.5663 + -34.8476*x + 0.9886*x^2});  
\addplot [>={Latex[length=1.5mm,width=0.5mm,angle'=25,open,round]},,domain=25:25.2,samples=2,line width=1.4128,color=blue]({129.7819 + 22.0355*x + -0.53808*x^2},{464.5663 + -34.8476*x + 0.9886*x^2});  
\addplot [>={Latex[length=1.5mm,width=0.5mm,angle'=25,open,round]},,domain=25.2:25.4,samples=2,line width=1.4174,color=blue]({129.7819 + 22.0355*x + -0.53808*x^2},{464.5663 + -34.8476*x + 0.9886*x^2});  
\addplot [>={Latex[length=1.5mm,width=0.5mm,angle'=25,open,round]},,domain=25.4:25.6,samples=3,line width=1.4147,color=blue]({129.7819 + 22.0355*x + -0.53808*x^2},{464.5663 + -34.8476*x + 0.9886*x^2});  
\addplot [>={Latex[length=1.5mm,width=0.5mm,angle'=25,open,round]},,domain=25.6:25.8,samples=2,line width=1.4073,color=blue]({129.7819 + 22.0355*x + -0.53808*x^2},{464.5663 + -34.8476*x + 0.9886*x^2});  
\addplot [>={Latex[length=1.5mm,width=0.5mm,angle'=25,open,round]},,domain=25.8:26,samples=2,line width=1.3965,color=blue]({129.7819 + 22.0355*x + -0.53808*x^2},{464.5663 + -34.8476*x + 0.9886*x^2});  
\addplot [>={Latex[length=1.5mm,width=0.5mm,angle'=25,open,round]},,domain=26:26.2,samples=2,line width=1.3826,color=blue]({129.7819 + 22.0355*x + -0.53808*x^2},{464.5663 + -34.8476*x + 0.9886*x^2});  
\addplot [>={Latex[length=1.5mm,width=0.5mm,angle'=25,open,round]},,domain=26.2:26.4,samples=2,line width=1.3663,color=blue]({129.7819 + 22.0355*x + -0.53808*x^2},{464.5663 + -34.8476*x + 0.9886*x^2});  
\addplot [>={Latex[length=1.5mm,width=0.5mm,angle'=25,open,round]},,domain=26.4:26.6,samples=3,line width=1.3489,color=blue]({129.7819 + 22.0355*x + -0.53808*x^2},{464.5663 + -34.8476*x + 0.9886*x^2});  
\addplot [>={Latex[length=1.5mm,width=0.5mm,angle'=25,open,round]},,domain=26.6:26.8,samples=2,line width=1.3296,color=blue]({129.7819 + 22.0355*x + -0.53808*x^2},{464.5663 + -34.8476*x + 0.9886*x^2});  
\addplot [>={Latex[length=1.5mm,width=0.5mm,angle'=25,open,round]},,domain=26.8:27,samples=2,line width=1.3096,color=blue]({129.7819 + 22.0355*x + -0.53808*x^2},{464.5663 + -34.8476*x + 0.9886*x^2});  
\addplot [>={Latex[length=1.5mm,width=0.5mm,angle'=25,open,round]},,domain=27:27.2,samples=2,line width=1.2924,color=blue]({129.7819 + 22.0355*x + -0.53808*x^2},{464.5663 + -34.8476*x + 0.9886*x^2});  
\addplot [>={Latex[length=1.5mm,width=0.5mm,angle'=25,open,round]},,domain=27.2:27.4,samples=2,line width=1.2791,color=blue]({129.7819 + 22.0355*x + -0.53808*x^2},{464.5663 + -34.8476*x + 0.9886*x^2});  
\addplot [>={Latex[length=1.5mm,width=0.5mm,angle'=25,open,round]},,domain=27.4:27.6,samples=3,line width=1.2687,color=blue]({129.7819 + 22.0355*x + -0.53808*x^2},{464.5663 + -34.8476*x + 0.9886*x^2});  
\addplot [>={Latex[length=1.5mm,width=0.5mm,angle'=25,open,round]},,domain=27.6:27.8,samples=2,line width=1.2609,color=blue]({129.7819 + 22.0355*x + -0.53808*x^2},{464.5663 + -34.8476*x + 0.9886*x^2});  
\addplot [>={Latex[length=1.5mm,width=0.5mm,angle'=25,open,round]},,domain=27.8:28,samples=2,line width=1.2558,color=blue]({129.7819 + 22.0355*x + -0.53808*x^2},{464.5663 + -34.8476*x + 0.9886*x^2});  
\addplot [>={Latex[length=1.5mm,width=0.5mm,angle'=25,open,round]},,domain=0:1,samples=2,line width=1.2558,color=blue]({324.9212 + 6.8212e-13*x},{263.8984 + 2.1032e-12*x});  
\addplot [>={Latex[length=1.5mm,width=0.5mm,angle'=25,open,round]},,domain=0:1,samples=2,line width=1.2558,color=blue]({324.9212 + -9.9171*x},{263.8984 + -5.9821*x});  
\addplot [>={Latex[length=1.5mm,width=0.5mm,angle'=25,open,round]},,domain=0:1,samples=2,line width=1.2558,color=blue]({315.0041 + 5.9742e-11*x},{257.9163 + -3.4152e-10*x});  
\addplot [>={Latex[length=1.5mm,width=0.5mm,angle'=25,open,round]},,domain=28:28.2,samples=2,line width=1.254,color=blue]({-266.56 + 40.0852*x + -0.68982*x^2},{1527.7863 + -78.2441*x + 1.1747*x^2});  
\addplot [>={Latex[length=1.5mm,width=0.5mm,angle'=25,open,round]},,domain=28.2:28.4,samples=2,line width=1.2547,color=blue]({-266.56 + 40.0852*x + -0.68982*x^2},{1527.7863 + -78.2441*x + 1.1747*x^2});  
\addplot [>={Latex[length=1.5mm,width=0.5mm,angle'=25,open,round]},,domain=28.4:28.6,samples=3,line width=1.2553,color=blue]({-266.56 + 40.0852*x + -0.68982*x^2},{1527.7863 + -78.2441*x + 1.1747*x^2});  
\addplot [>={Latex[length=1.5mm,width=0.5mm,angle'=25,open,round]},,domain=28.6:28.8,samples=2,line width=1.2598,color=blue]({-266.56 + 40.0852*x + -0.68982*x^2},{1527.7863 + -78.2441*x + 1.1747*x^2});  
\addplot [>={Latex[length=1.5mm,width=0.5mm,angle'=25,open,round]},,domain=28.8:29,samples=2,line width=1.2643,color=blue]({-266.56 + 40.0852*x + -0.68982*x^2},{1527.7863 + -78.2441*x + 1.1747*x^2});  
\addplot [>={Latex[length=1.5mm,width=0.5mm,angle'=25,open,round]},,domain=29:29.2,samples=2,line width=1.2709,color=blue]({-266.56 + 40.0852*x + -0.68982*x^2},{1527.7863 + -78.2441*x + 1.1747*x^2});  
\addplot [>={Latex[length=1.5mm,width=0.5mm,angle'=25,open,round]},,domain=29.2:29.4,samples=2,line width=1.2795,color=blue]({-266.56 + 40.0852*x + -0.68982*x^2},{1527.7863 + -78.2441*x + 1.1747*x^2});  
\addplot [>={Latex[length=1.5mm,width=0.5mm,angle'=25,open,round]},,domain=29.4:29.6,samples=3,line width=1.2882,color=blue]({-266.56 + 40.0852*x + -0.68982*x^2},{1527.7863 + -78.2441*x + 1.1747*x^2});  
\addplot [>={Latex[length=1.5mm,width=0.5mm,angle'=25,open,round]},,domain=29.6:29.8,samples=2,line width=1.2971,color=blue]({-266.56 + 40.0852*x + -0.68982*x^2},{1527.7863 + -78.2441*x + 1.1747*x^2});  
\addplot [>={Latex[length=1.5mm,width=0.5mm,angle'=25,open,round]},,domain=29.8:30,samples=2,line width=1.306,color=blue]({-266.56 + 40.0852*x + -0.68982*x^2},{1527.7863 + -78.2441*x + 1.1747*x^2});  
\addplot [>={Latex[length=1.5mm,width=0.5mm,angle'=25,open,round]},,domain=30:30.2,samples=2,line width=1.3124,color=blue]({-266.56 + 40.0852*x + -0.68982*x^2},{1527.7863 + -78.2441*x + 1.1747*x^2});  
\addplot [>={Latex[length=1.5mm,width=0.5mm,angle'=25,open,round]},,domain=30.2:30.4,samples=2,line width=1.3164,color=blue]({-266.56 + 40.0852*x + -0.68982*x^2},{1527.7863 + -78.2441*x + 1.1747*x^2});  
\addplot [>={Latex[length=1.5mm,width=0.5mm,angle'=25,open,round]},,domain=30.4:30.6,samples=3,line width=1.3203,color=blue]({-266.56 + 40.0852*x + -0.68982*x^2},{1527.7863 + -78.2441*x + 1.1747*x^2});  
\addplot [>={Latex[length=1.5mm,width=0.5mm,angle'=25,open,round]},,domain=30.6:30.8,samples=2,line width=1.3243,color=blue]({-266.56 + 40.0852*x + -0.68982*x^2},{1527.7863 + -78.2441*x + 1.1747*x^2});  
\addplot [>={Latex[length=1.5mm,width=0.5mm,angle'=25,open,round]},,domain=30.8:31,samples=2,line width=1.3282,color=blue]({-266.56 + 40.0852*x + -0.68982*x^2},{1527.7863 + -78.2441*x + 1.1747*x^2});  
\addplot [>={Latex[length=1.5mm,width=0.5mm,angle'=25,open,round]},,domain=31:31.2,samples=2,line width=1.3322,color=blue]({-266.56 + 40.0852*x + -0.68982*x^2},{1527.7863 + -78.2441*x + 1.1747*x^2});  
\addplot [>={Latex[length=1.5mm,width=0.5mm,angle'=25,open,round]},,domain=31.2:31.4,samples=2,line width=1.3362,color=blue]({-266.56 + 40.0852*x + -0.68982*x^2},{1527.7863 + -78.2441*x + 1.1747*x^2});  
\addplot [>={Latex[length=1.5mm,width=0.5mm,angle'=25,open,round]},,domain=31.4:31.6,samples=3,line width=1.3402,color=blue]({-266.56 + 40.0852*x + -0.68982*x^2},{1527.7863 + -78.2441*x + 1.1747*x^2});  
\addplot [>={Latex[length=1.5mm,width=0.5mm,angle'=25,open,round]},,domain=31.6:31.8,samples=2,line width=1.3442,color=blue]({-266.56 + 40.0852*x + -0.68982*x^2},{1527.7863 + -78.2441*x + 1.1747*x^2});  
\addplot [>={Latex[length=1.5mm,width=0.5mm,angle'=25,open,round]},,domain=31.8:32,samples=2,line width=1.3482,color=blue]({-266.56 + 40.0852*x + -0.68982*x^2},{1527.7863 + -78.2441*x + 1.1747*x^2});  
\addplot [>={Latex[length=1.5mm,width=0.5mm,angle'=25,open,round]},,domain=32:32.2,samples=3,line width=1.3511,color=blue]({-266.56 + 40.0852*x + -0.68982*x^2},{1527.7863 + -78.2441*x + 1.1747*x^2});  
\addplot [>={Latex[length=1.5mm,width=0.5mm,angle'=25,open,round]},,domain=32.2:32.4,samples=2,line width=1.353,color=blue]({-266.56 + 40.0852*x + -0.68982*x^2},{1527.7863 + -78.2441*x + 1.1747*x^2});  
\addplot [>={Latex[length=1.5mm,width=0.5mm,angle'=25,open,round]},,domain=32.4:32.6,samples=3,line width=1.3548,color=blue]({-266.56 + 40.0852*x + -0.68982*x^2},{1527.7863 + -78.2441*x + 1.1747*x^2});  
\addplot [>={Latex[length=1.5mm,width=0.5mm,angle'=25,open,round]},,domain=32.6:32.8,samples=2,line width=1.3567,color=blue]({-266.56 + 40.0852*x + -0.68982*x^2},{1527.7863 + -78.2441*x + 1.1747*x^2});  
\addplot [>={Latex[length=1.5mm,width=0.5mm,angle'=25,open,round]},,domain=32.8:33,samples=3,line width=1.3586,color=blue]({-266.56 + 40.0852*x + -0.68982*x^2},{1527.7863 + -78.2441*x + 1.1747*x^2});  
\addplot [>={Latex[length=1.5mm,width=0.5mm,angle'=25,open,round]},,domain=33:33.2,samples=3,line width=1.3602,color=blue]({-266.56 + 40.0852*x + -0.68982*x^2},{1527.7863 + -78.2441*x + 1.1747*x^2});  
\addplot [>={Latex[length=1.5mm,width=0.5mm,angle'=25,open,round]},,domain=33.2:33.4,samples=2,line width=1.3617,color=blue]({-266.56 + 40.0852*x + -0.68982*x^2},{1527.7863 + -78.2441*x + 1.1747*x^2});  
\addplot [>={Latex[length=1.5mm,width=0.5mm,angle'=25,open,round]},,domain=33.4:33.6,samples=3,line width=1.3631,color=blue]({-266.56 + 40.0852*x + -0.68982*x^2},{1527.7863 + -78.2441*x + 1.1747*x^2});  
\addplot [>={Latex[length=1.5mm,width=0.5mm,angle'=25,open,round]},,domain=33.6:33.8,samples=2,line width=1.3646,color=blue]({-266.56 + 40.0852*x + -0.68982*x^2},{1527.7863 + -78.2441*x + 1.1747*x^2});  
\addplot [>={Latex[length=1.5mm,width=0.5mm,angle'=25,open,round]},,domain=33.8:34,samples=3,line width=1.366,color=blue]({-266.56 + 40.0852*x + -0.68982*x^2},{1527.7863 + -78.2441*x + 1.1747*x^2});  
\addplot [>={Latex[length=1.5mm,width=0.5mm,angle'=25,open,round]},,domain=0:1,samples=2,line width=1.366,color=blue]({298.9009 + -3.4277e-11*x},{225.44 + 1.9639e-10*x});  
\addplot [>={Latex[length=1.5mm,width=0.5mm,angle'=25,open,round]},,domain=0:1,samples=2,line width=1.366,color=blue]({298.9009 + -15.5916*x},{225.44 + 6.3574*x});  
\addplot [->,>={Latex[length=1.5mm,width=0.5mm,angle'=25,open,round]},,domain=0:1,samples=2,line width=1.366,color=blue]({283.3093 + 3.979e-13*x},{231.7974 + -4.9738e-12*x});  
\addplot [->,>={Latex[length=1.5mm,width=0.5mm,angle'=25,open,round]},,domain=34:34.2,samples=3,line width=1.3675,color=blue]({13.9503 + 15.9009*x + -0.23466*x^2},{-176.1536 + 16.2206*x + -0.12418*x^2});  
\addplot [->,>={Latex[length=1.5mm,width=0.5mm,angle'=25,open,round]},,domain=34.2:34.4,samples=2,line width=1.3691,color=blue]({13.9503 + 15.9009*x + -0.23466*x^2},{-176.1536 + 16.2206*x + -0.12418*x^2});  
\addplot [->,>={Latex[length=1.5mm,width=0.5mm,angle'=25,open,round]},,domain=34.4:34.6,samples=3,line width=1.3707,color=blue]({13.9503 + 15.9009*x + -0.23466*x^2},{-176.1536 + 16.2206*x + -0.12418*x^2});  
\addplot [->,>={Latex[length=1.5mm,width=0.5mm,angle'=25,open,round]},,domain=34.6:34.8,samples=2,line width=1.3722,color=blue]({13.9503 + 15.9009*x + -0.23466*x^2},{-176.1536 + 16.2206*x + -0.12418*x^2});  
\addplot [->,>={Latex[length=1.5mm,width=0.5mm,angle'=25,open,round]},,domain=34.8:35,samples=3,line width=1.3738,color=blue]({13.9503 + 15.9009*x + -0.23466*x^2},{-176.1536 + 16.2206*x + -0.12418*x^2});  
\addplot [->,>={Latex[length=1.5mm,width=0.5mm,angle'=25,open,round]},,domain=35:35.2,samples=3,line width=1.3876,color=blue]({13.9503 + 15.9009*x + -0.23466*x^2},{-176.1536 + 16.2206*x + -0.12418*x^2});  
\addplot [->,>={Latex[length=1.5mm,width=0.5mm,angle'=25,open,round]},,domain=35.2:35.4,samples=2,line width=1.4136,color=blue]({13.9503 + 15.9009*x + -0.23466*x^2},{-176.1536 + 16.2206*x + -0.12418*x^2});  
\addplot [->,>={Latex[length=1.5mm,width=0.5mm,angle'=25,open,round]},,domain=35.4:35.6,samples=3,line width=1.4397,color=blue]({13.9503 + 15.9009*x + -0.23466*x^2},{-176.1536 + 16.2206*x + -0.12418*x^2});  
\addplot [->,>={Latex[length=1.5mm,width=0.5mm,angle'=25,open,round]},,domain=35.6:35.8,samples=2,line width=1.4657,color=blue]({13.9503 + 15.9009*x + -0.23466*x^2},{-176.1536 + 16.2206*x + -0.12418*x^2});  
\addplot [->,>={Latex[length=1.5mm,width=0.5mm,angle'=25,open,round]},,domain=35.8:36,samples=3,line width=1.4918,color=blue]({13.9503 + 15.9009*x + -0.23466*x^2},{-176.1536 + 16.2206*x + -0.12418*x^2});  
\addplot [->,>={Latex[length=1.5mm,width=0.5mm,angle'=25,open,round]},,domain=36:36.2,samples=3,line width=1.515,color=blue]({13.9503 + 15.9009*x + -0.23466*x^2},{-176.1536 + 16.2206*x + -0.12418*x^2});  
\addplot [->,>={Latex[length=1.5mm,width=0.5mm,angle'=25,open,round]},,domain=36.2:36.4,samples=2,line width=1.5353,color=blue]({13.9503 + 15.9009*x + -0.23466*x^2},{-176.1536 + 16.2206*x + -0.12418*x^2});  
\addplot [->,>={Latex[length=1.5mm,width=0.5mm,angle'=25,open,round]},,domain=36.4:36.6,samples=3,line width=1.5556,color=blue]({13.9503 + 15.9009*x + -0.23466*x^2},{-176.1536 + 16.2206*x + -0.12418*x^2});  
\addplot [->,>={Latex[length=1.5mm,width=0.5mm,angle'=25,open,round]},,domain=36.6:36.8,samples=2,line width=1.576,color=blue]({13.9503 + 15.9009*x + -0.23466*x^2},{-176.1536 + 16.2206*x + -0.12418*x^2});  
\addplot [->,>={Latex[length=1.5mm,width=0.5mm,angle'=25,open,round]},,domain=36.8:37,samples=3,line width=1.5963,color=blue]({13.9503 + 15.9009*x + -0.23466*x^2},{-176.1536 + 16.2206*x + -0.12418*x^2});  
\addplot [->,>={Latex[length=1.5mm,width=0.5mm,angle'=25,open,round]},,domain=37:37.2,samples=3,line width=1.6181,color=blue]({13.9503 + 15.9009*x + -0.23466*x^2},{-176.1536 + 16.2206*x + -0.12418*x^2});  
\addplot [->,>={Latex[length=1.5mm,width=0.5mm,angle'=25,open,round]},,domain=37.2:37.4,samples=2,line width=1.6413,color=blue]({13.9503 + 15.9009*x + -0.23466*x^2},{-176.1536 + 16.2206*x + -0.12418*x^2});  
\addplot [->,>={Latex[length=1.5mm,width=0.5mm,angle'=25,open,round]},,domain=37.4:37.6,samples=3,line width=1.6646,color=blue]({13.9503 + 15.9009*x + -0.23466*x^2},{-176.1536 + 16.2206*x + -0.12418*x^2});  
\addplot [->,>={Latex[length=1.5mm,width=0.5mm,angle'=25,open,round]},,domain=37.6:37.8,samples=2,line width=1.6878,color=blue]({13.9503 + 15.9009*x + -0.23466*x^2},{-176.1536 + 16.2206*x + -0.12418*x^2});  
\addplot [->,>={Latex[length=1.5mm,width=0.5mm,angle'=25,open,round]},,domain=37.8:38,samples=3,line width=1.711,color=blue]({13.9503 + 15.9009*x + -0.23466*x^2},{-176.1536 + 16.2206*x + -0.12418*x^2});  
\addplot [->,>={Latex[length=1.5mm,width=0.5mm,angle'=25,open,round]},,domain=38:38.2,samples=3,line width=1.7343,color=blue]({13.9503 + 15.9009*x + -0.23466*x^2},{-176.1536 + 16.2206*x + -0.12418*x^2});  
\addplot [->,>={Latex[length=1.5mm,width=0.5mm,angle'=25,open,round]},,domain=38.2:38.4,samples=2,line width=1.7576,color=blue]({13.9503 + 15.9009*x + -0.23466*x^2},{-176.1536 + 16.2206*x + -0.12418*x^2});  
\addplot [->,>={Latex[length=1.5mm,width=0.5mm,angle'=25,open,round]},,domain=38.4:38.6,samples=3,line width=1.781,color=blue]({13.9503 + 15.9009*x + -0.23466*x^2},{-176.1536 + 16.2206*x + -0.12418*x^2});  
\addplot [->,>={Latex[length=1.5mm,width=0.5mm,angle'=25,open,round]},,domain=38.6:38.8,samples=2,line width=1.8043,color=blue]({13.9503 + 15.9009*x + -0.23466*x^2},{-176.1536 + 16.2206*x + -0.12418*x^2});  
\addplot [->,>={Latex[length=1.5mm,width=0.5mm,angle'=25,open,round]},,domain=38.8:39,samples=3,line width=1.8277,color=blue]({13.9503 + 15.9009*x + -0.23466*x^2},{-176.1536 + 16.2206*x + -0.12418*x^2});  
\addplot [->,>={Latex[length=1.5mm,width=0.5mm,angle'=25,open,round]},,domain=39:39.2,samples=3,line width=1.8512,color=blue]({13.9503 + 15.9009*x + -0.23466*x^2},{-176.1536 + 16.2206*x + -0.12418*x^2});  
\addplot [->,>={Latex[length=1.5mm,width=0.5mm,angle'=25,open,round]},,domain=39.2:39.4,samples=2,line width=1.8748,color=blue]({13.9503 + 15.9009*x + -0.23466*x^2},{-176.1536 + 16.2206*x + -0.12418*x^2});  
\addplot [->,>={Latex[length=1.5mm,width=0.5mm,angle'=25,open,round]},,domain=39.4:39.6,samples=3,line width=1.8984,color=blue]({13.9503 + 15.9009*x + -0.23466*x^2},{-176.1536 + 16.2206*x + -0.12418*x^2});  
\addplot [->,>={Latex[length=1.5mm,width=0.5mm,angle'=25,open,round]},,domain=39.6:39.8,samples=2,line width=1.9221,color=blue]({13.9503 + 15.9009*x + -0.23466*x^2},{-176.1536 + 16.2206*x + -0.12418*x^2});  
\addplot [->,>={Latex[length=1.5mm,width=0.5mm,angle'=25,open,round]},,domain=39.8:40,samples=3,line width=1.9457,color=blue]({13.9503 + 15.9009*x + -0.23466*x^2},{-176.1536 + 16.2206*x + -0.12418*x^2});  
\addplot [->,>={Latex[length=1.5mm,width=0.5mm,angle'=25,open,round]},,domain=40:40.2,samples=3,line width=1.9694,color=blue]({13.9503 + 15.9009*x + -0.23466*x^2},{-176.1536 + 16.2206*x + -0.12418*x^2});  
\addplot [->,>={Latex[length=1.5mm,width=0.5mm,angle'=25,open,round]},,domain=40.2:40.4,samples=2,line width=1.9931,color=blue]({13.9503 + 15.9009*x + -0.23466*x^2},{-176.1536 + 16.2206*x + -0.12418*x^2});  
\addplot [->,>={Latex[length=1.5mm,width=0.5mm,angle'=25,open,round]},,domain=40.4:40.6,samples=3,line width=2.0168,color=blue]({13.9503 + 15.9009*x + -0.23466*x^2},{-176.1536 + 16.2206*x + -0.12418*x^2});  
\addplot [->,>={Latex[length=1.5mm,width=0.5mm,angle'=25,open,round]},,domain=40.6:40.8,samples=2,line width=2.0405,color=blue]({13.9503 + 15.9009*x + -0.23466*x^2},{-176.1536 + 16.2206*x + -0.12418*x^2});  
\addplot [->,>={Latex[length=1.5mm,width=0.5mm,angle'=25,open,round]},,domain=40.8:41,samples=3,line width=2.0643,color=blue]({13.9503 + 15.9009*x + -0.23466*x^2},{-176.1536 + 16.2206*x + -0.12418*x^2});  
\addplot [->,>={Latex[length=1.5mm,width=0.5mm,angle'=25,open,round]},,domain=41:41.2,samples=3,line width=2.0875,color=blue]({13.9503 + 15.9009*x + -0.23466*x^2},{-176.1536 + 16.2206*x + -0.12418*x^2});  
\addplot [->,>={Latex[length=1.5mm,width=0.5mm,angle'=25,open,round]},,domain=41.2:41.4,samples=2,line width=2.1104,color=blue]({13.9503 + 15.9009*x + -0.23466*x^2},{-176.1536 + 16.2206*x + -0.12418*x^2});  
\addplot [->,>={Latex[length=1.5mm,width=0.5mm,angle'=25,open,round]},,domain=41.4:41.6,samples=3,line width=2.1332,color=blue]({13.9503 + 15.9009*x + -0.23466*x^2},{-176.1536 + 16.2206*x + -0.12418*x^2});  
\addplot [->,>={Latex[length=1.5mm,width=0.5mm,angle'=25,open,round]},,domain=41.6:41.8,samples=2,line width=2.156,color=blue]({13.9503 + 15.9009*x + -0.23466*x^2},{-176.1536 + 16.2206*x + -0.12418*x^2});  
\addplot [->,>={Latex[length=1.5mm,width=0.5mm,angle'=25,open,round]},,domain=41.8:42,samples=3,line width=2.1789,color=blue]({13.9503 + 15.9009*x + -0.23466*x^2},{-176.1536 + 16.2206*x + -0.12418*x^2});  
\addplot [->,>={Latex[length=1.5mm,width=0.5mm,angle'=25,open,round]},,domain=42:42.2,samples=3,line width=2.2025,color=blue]({13.9503 + 15.9009*x + -0.23466*x^2},{-176.1536 + 16.2206*x + -0.12418*x^2});  
\addplot [->,>={Latex[length=1.5mm,width=0.5mm,angle'=25,open,round]},,domain=42.2:42.4,samples=2,line width=2.227,color=blue]({13.9503 + 15.9009*x + -0.23466*x^2},{-176.1536 + 16.2206*x + -0.12418*x^2});  
\addplot [->,>={Latex[length=1.5mm,width=0.5mm,angle'=25,open,round]},,domain=42.4:42.6,samples=3,line width=2.2515,color=blue]({13.9503 + 15.9009*x + -0.23466*x^2},{-176.1536 + 16.2206*x + -0.12418*x^2});  
\addplot [->,>={Latex[length=1.5mm,width=0.5mm,angle'=25,open,round]},,domain=42.6:42.8,samples=2,line width=2.276,color=blue]({13.9503 + 15.9009*x + -0.23466*x^2},{-176.1536 + 16.2206*x + -0.12418*x^2});  
\addplot [->,>={Latex[length=1.5mm,width=0.5mm,angle'=25,open,round]},,domain=42.8:43,samples=3,line width=2.3005,color=blue]({13.9503 + 15.9009*x + -0.23466*x^2},{-176.1536 + 16.2206*x + -0.12418*x^2});  
\addplot [->,>={Latex[length=1.5mm,width=0.5mm,angle'=25,open,round]},,domain=43:43.2,samples=3,line width=2.3212,color=blue]({13.9503 + 15.9009*x + -0.23466*x^2},{-176.1536 + 16.2206*x + -0.12418*x^2});  
\addplot [->,>={Latex[length=1.5mm,width=0.5mm,angle'=25,open,round]},,domain=43.2:43.4,samples=2,line width=2.338,color=blue]({13.9503 + 15.9009*x + -0.23466*x^2},{-176.1536 + 16.2206*x + -0.12418*x^2});  
\addplot [->,>={Latex[length=1.5mm,width=0.5mm,angle'=25,open,round]},,domain=43.4:43.6,samples=3,line width=2.3548,color=blue]({13.9503 + 15.9009*x + -0.23466*x^2},{-176.1536 + 16.2206*x + -0.12418*x^2});  
\addplot [->,>={Latex[length=1.5mm,width=0.5mm,angle'=25,open,round]},,domain=43.6:43.8,samples=2,line width=2.3715,color=blue]({13.9503 + 15.9009*x + -0.23466*x^2},{-176.1536 + 16.2206*x + -0.12418*x^2});  
\addplot [->,>={Latex[length=1.5mm,width=0.5mm,angle'=25,open,round]},,domain=43.8:44,samples=3,line width=2.3883,color=blue]({13.9503 + 15.9009*x + -0.23466*x^2},{-176.1536 + 16.2206*x + -0.12418*x^2});  
\addplot [->,>={Latex[length=1.5mm,width=0.5mm,angle'=25,open,round]},,domain=44:44.2,samples=3,line width=2.4051,color=blue]({13.9503 + 15.9009*x + -0.23466*x^2},{-176.1536 + 16.2206*x + -0.12418*x^2});  
\addplot [->,>={Latex[length=1.5mm,width=0.5mm,angle'=25,open,round]},,domain=44.2:44.4,samples=2,line width=2.4219,color=blue]({13.9503 + 15.9009*x + -0.23466*x^2},{-176.1536 + 16.2206*x + -0.12418*x^2});  
\addplot [->,>={Latex[length=1.5mm,width=0.5mm,angle'=25,open,round]},,domain=44.4:44.6,samples=3,line width=2.4386,color=blue]({13.9503 + 15.9009*x + -0.23466*x^2},{-176.1536 + 16.2206*x + -0.12418*x^2});  
\addplot [->,>={Latex[length=1.5mm,width=0.5mm,angle'=25,open,round]},,domain=44.6:44.8,samples=2,line width=2.4554,color=blue]({13.9503 + 15.9009*x + -0.23466*x^2},{-176.1536 + 16.2206*x + -0.12418*x^2});  
\addplot [->,>={Latex[length=1.5mm,width=0.5mm,angle'=25,open,round]},,domain=44.8:45,samples=3,line width=2.4722,color=blue]({13.9503 + 15.9009*x + -0.23466*x^2},{-176.1536 + 16.2206*x + -0.12418*x^2});  
\addplot [->,>={Latex[length=1.5mm,width=0.5mm,angle'=25,open,round]},,domain=45:45.2,samples=3,line width=2.4889,color=blue]({13.9503 + 15.9009*x + -0.23466*x^2},{-176.1536 + 16.2206*x + -0.12418*x^2});  
\addplot [->,>={Latex[length=1.5mm,width=0.5mm,angle'=25,open,round]},,domain=45.2:45.4,samples=2,line width=2.5,color=blue]({13.9503 + 15.9009*x + -0.23466*x^2},{-176.1536 + 16.2206*x + -0.12418*x^2});  
\addplot [->,>={Latex[length=1.5mm,width=0.5mm,angle'=25,open,round]},,domain=45.4:45.6,samples=3,line width=2.5,color=blue]({13.9503 + 15.9009*x + -0.23466*x^2},{-176.1536 + 16.2206*x + -0.12418*x^2});  
\addplot [->,>={Latex[length=1.5mm,width=0.5mm,angle'=25,open,round]},,domain=45.6:45.8,samples=2,line width=2.5,color=blue]({13.9503 + 15.9009*x + -0.23466*x^2},{-176.1536 + 16.2206*x + -0.12418*x^2});  
\addplot [->,>={Latex[length=1.5mm,width=0.5mm,angle'=25,open,round]},,domain=45.8:46,samples=3,line width=2.5,color=blue]({13.9503 + 15.9009*x + -0.23466*x^2},{-176.1536 + 16.2206*x + -0.12418*x^2});  
\addplot [->,>={Latex[length=1.5mm,width=0.5mm,angle'=25,open,round]},,domain=46:46.2,samples=3,line width=2.5,color=blue]({13.9503 + 15.9009*x + -0.23466*x^2},{-176.1536 + 16.2206*x + -0.12418*x^2});  
\addplot [->,>={Latex[length=1.5mm,width=0.5mm,angle'=25,open,round]},,domain=46.2:46.4,samples=2,line width=2.5,color=blue]({13.9503 + 15.9009*x + -0.23466*x^2},{-176.1536 + 16.2206*x + -0.12418*x^2});  
\addplot [->,>={Latex[length=1.5mm,width=0.5mm,angle'=25,open,round]},,domain=46.4:46.6,samples=3,line width=2.5,color=blue]({13.9503 + 15.9009*x + -0.23466*x^2},{-176.1536 + 16.2206*x + -0.12418*x^2});  
\addplot [->,>={Latex[length=1.5mm,width=0.5mm,angle'=25,open,round]},,domain=46.6:46.8,samples=2,line width=2.5,color=blue]({13.9503 + 15.9009*x + -0.23466*x^2},{-176.1536 + 16.2206*x + -0.12418*x^2});  
\addplot [->,>={Latex[length=1.5mm,width=0.5mm,angle'=25,open,round]},,domain=46.8:47,samples=3,line width=2.5,color=blue]({13.9503 + 15.9009*x + -0.23466*x^2},{-176.1536 + 16.2206*x + -0.12418*x^2});  
\addplot [->,>={Latex[length=1.5mm,width=0.5mm,angle'=25,open,round]},,domain=47:47.2,samples=3,line width=2.5,color=blue]({13.9503 + 15.9009*x + -0.23466*x^2},{-176.1536 + 16.2206*x + -0.12418*x^2});  
\addplot [->,>={Latex[length=1.5mm,width=0.5mm,angle'=25,open,round]},,domain=47.2:47.4,samples=2,line width=2.5,color=blue]({13.9503 + 15.9009*x + -0.23466*x^2},{-176.1536 + 16.2206*x + -0.12418*x^2});  
\addplot [->,>={Latex[length=1.5mm,width=0.5mm,angle'=25,open,round]},,domain=47.4:47.6,samples=3,line width=2.5,color=blue]({13.9503 + 15.9009*x + -0.23466*x^2},{-176.1536 + 16.2206*x + -0.12418*x^2});  
\addplot [->,>={Latex[length=1.5mm,width=0.5mm,angle'=25,open,round]},,domain=47.6:47.8,samples=2,line width=2.5,color=blue]({13.9503 + 15.9009*x + -0.23466*x^2},{-176.1536 + 16.2206*x + -0.12418*x^2});  
\addplot [->,>={Latex[length=1.5mm,width=0.5mm,angle'=25,open,round]},,domain=47.8:48,samples=3,line width=2.5,color=blue]({13.9503 + 15.9009*x + -0.23466*x^2},{-176.1536 + 16.2206*x + -0.12418*x^2});  
\addplot [->,>={Latex[length=1.5mm,width=0.5mm,angle'=25,open,round]},,domain=48:48.2,samples=3,line width=2.5,color=blue]({13.9503 + 15.9009*x + -0.23466*x^2},{-176.1536 + 16.2206*x + -0.12418*x^2});  
\addplot [->,>={Latex[length=1.5mm,width=0.5mm,angle'=25,open,round]},,domain=48.2:48.4,samples=2,line width=2.5,color=blue]({13.9503 + 15.9009*x + -0.23466*x^2},{-176.1536 + 16.2206*x + -0.12418*x^2});  
\addplot [->,>={Latex[length=1.5mm,width=0.5mm,angle'=25,open,round]},,domain=48.4:48.6,samples=3,line width=2.5,color=blue]({13.9503 + 15.9009*x + -0.23466*x^2},{-176.1536 + 16.2206*x + -0.12418*x^2});  
\addplot [->,>={Latex[length=1.5mm,width=0.5mm,angle'=25,open,round]},,domain=48.6:48.8,samples=2,line width=2.5,color=blue]({13.9503 + 15.9009*x + -0.23466*x^2},{-176.1536 + 16.2206*x + -0.12418*x^2});  
\addplot [->,>={Latex[length=1.5mm,width=0.5mm,angle'=25,open,round]},,domain=48.8:49,samples=3,line width=2.5,color=blue]({13.9503 + 15.9009*x + -0.23466*x^2},{-176.1536 + 16.2206*x + -0.12418*x^2});  
\end{axis} 
\end{tikzpicture} 

  \noindent}

%% file: sections/method.tex
We propose the following pipeline to reconstruct a 6DoF pose of a fast moving object:
\begin{enumerate}
\item From a given 2D trajectory, in our case computed by the TbD-NC algorithm~\cite{tbdnc}, reconstruct sub-frame blur-free snapshots of the object by piece-wise deblatting (Section~\ref{sec:pw}).
\item Estimate the relative distance from the object to the camera from the estimated object shape mask (Section~\ref{sec:3d}).
\item Using the assumption of a spherical object with locally constant rotation find the rotation axis and velocity by minimizing the reprojection error (Section~\ref{sec:rot}).
\end{enumerate}

An alternative method to estimate the 3D rotation of fast moving objects from their snapshots would be to run a classical 3D reconstruction pipeline such as COLMAP~\cite{colmap}. We tried reconstruction and structure-from-motion pipelines~\cite{colmap,Lourakis:2009:SSP:1486525.1486527,Hartley:2003:MVG:861369,cmpmvs} and none of them were able to deal with small objects containing few features. They do not perform well even on snapshots from a high-speed camera sequence, where the motion blur is negligible. 

Tracking by Deblatting in 3D thus extends TbD and TbD-NC by using trajectories estimated by these methods to infer more attributes about the object and its motion. The core of TbD consists of two alternating optimization steps. The first step updates the object shape and appearance $(F,M)$ while the trajectory $H$ is fixed, and the second one updates the trajectory $H$ while the object $(F,M)$ is fixed. Both steps are formulated as convex optimization problems with non-smooth terms and constraints and solved using the ADMM method \cite{Boyd2011}. Throughout processing of the video sequence, TbD maintains a long-term appearance model $\hat F$ that is used to regularize the estimation of $F$ in the new frame.

We have made two modifications to the TbD core steps. First, we added a new regularization term to the shape-and-appearance $(F,M)$ estimation step to facilitate shape estimation in cases when the tracked object is a ball and its shape is thus circularly symmetric. The modified optimization problem is
\begin{multline}
\label{eq:FM_loss}
\min_{F,M}\frac{1}{2}\left\|H*F+(1-H*M)B-I\right\|_2^2\\
+\frac{\lambda}{2}\|F-\hat F\|_2^2	+\alpha_F\|\nabla F\|_1 +\frac{\lambda_R}{2}\|\mathbf{R}M-M\|_2^2,
\end{multline}
s.t. $0\leq F\leq M\leq 1$, where matrix inequalities are considered element-wise. The first term is the data likelihood given by the image formation model \eqref{eq:acquisition_model}. The second term constrains the solution to be close to the template $\hat F$ and the third term is Total Variation that enforces piece-wise smooth object appearance. In the last, $\lambda_R$-weighted term, $\mathbf{R}$ is a linear operator that performs circular averaging, i.e. the shape mask $M$ is forced to be rotationally symmetric.

Second, in the estimation of $H$ we replaced the $L^1$ regularization of $H$ by the constraint $\sum_i H[i]=1$, which is free of weighting parameters that have to be tuned for different sequences. The modified optimization problem is then
\begin{equation}
\label{eq:deconv_H_loss}
\begin{gathered}
\min_H\frac{1}{2}\left\|H*F+(1-H*M)B-I\right\|_2^2,\\
\text{s.t. }H \geq 0,\,\sum_i H[i]=1.
\end{gathered}
\end{equation}

\input{tables/comparison.tex}

\subsection{Piece-wise Deblatting} 
\label{sec:pw}
TbD assumes that the appearance and shape of the object is constant during single frame exposure. In reality, the appearance changes even within a single video frame due to the object rotation and camera projection. We propose to approximately model this gradual change as a sequence of constant snapshots which we estimate. The snapshots can be used for temporal super-resolution and also to determine the intra-frame rotation of the object.

Suppose that the object trajectory in the form of a parametric curve $\mathcal{C}:[0,1]\to\mathbb{R}^2$ has been estimated for a given video frame. We partition this curve to multiple contiguous segments $\mathcal{C}_i$ with their corresponding blurs denoted $H_i$ and estimate the appearance and shape $(F_i,M_i)$ of the object at the time interval corresponding to $\mathcal{C}_i$. From the piecewise constant appearance assumption, we get the formation model of the video frame $I$ as 
\begin{equation}
	I = \sum_i H_i*F_i + \left(1-\sum\nolimits_i H_i*M_i\right)\cdot B.
\end{equation}
The optimization problem \eqref{eq:FM_loss} for joint estimation of $(F_i,M_i)$ on segments $\mathcal{C}_i$ becomes
\begin{multline}
	\label{eq:FM_loss_pw}
	\min_{F_i,M_i}\frac{1}{2}\left\|\sum\nolimits_i H_i*F_i+(1-\sum\nolimits_i H_i*M_i)B-I\right\|_2^2\\
	+\frac{\lambda}{2}\|F_i-\hat F_i\|_2^2+\alpha_F\|\nabla F\|_1+\frac{\lambda_R}{2}\sum_i\|\mathbf{R}M_i-M_i\|_2^2\\
	+\gamma_F\sum_i\|F_i-F_{i+1}\|_1+\gamma_M\sum_i\|M_i-M_{i+1}\|_1,
\end{multline}
s.t. $0\leq F_i\leq M_i\leq 1$.

The last two terms, weighted by $\gamma_{F}$ and $\gamma_{M}$, are new regularization terms enforcing similarity of both appearance and shape of the object in neighboring time intervals. $\hat F_i$ is a sub-frame extension of the appearance template used in TbD, regularizing the appearance estimation in corresponding segments.

The piecewise appearance estimation is implemented in a hierarchical manner. First, we split $\mathcal{C}$ into two segments $\mathcal{C}^1_1$ and $\mathcal{C}^1_2$ (superscript denotes the hierarchy level) and solve \eqref{eq:FM_loss_pw} for $F^1_{1}, F^1_{2}$ with both templates $\hat F^1_1 = \hat F^1_2 :=F^0$ where $F^0$ is the initial result of TbD. On the next level, we do another binary splitting of $\mathcal{C}^1_1$ to $\mathcal{C}^2_{1}, \mathcal{C}^2_{2}$ and $\mathcal{C}^1_2$ to $\mathcal{C}^2_{3}, \mathcal{C}^2_{4}$ and again solve \eqref{eq:FM_loss_pw} with templates set to results from the previous level, $\hat F^2_1 = \hat F^2_2 := F^1_1$ and $\hat F^2_3 = \hat F^2_4 :=F^1_2$. This process continues until the desirable number of splitting of $\mathcal{C}$ is achieved. Results of this estimation process are illustrated in Figure~\ref{fig:recon}.

\input{figures/perf_tiou.tex}

\subsection{3D Trajectory}
\label{sec:3d}
TbD-NC~\cite{tbdnc} provides a 2D part of the estimated trajectory by fitting piece-wise polynomial curves. We extend this approach to fitting piece-wise polynomial curve in 3D, where the third dimension is the object distance to the camera. We assume that the object is approximately spherical with radius $r$,~\ie the area of mask defined as sum of all pixel values is $\text{area}(M) := \sum_i M[i] = \pi r^2$.
The distance $d$ is inversely proportional to the perceived object radius $r$ and is given by
\begin{equation}
\label{eq:area}
d \propto \sqrt{\frac{\pi}{\text{area}(M)}}\,.
\end{equation}
Note that the absolute distance can be calculated if we know camera parameters and the actual object radius.
The estimated relative distances in sub-frame precision are expressed analytically by piece-wise continuous curve fitting. First, bounces are found as the ones initially estimated in 2D trajectory and then additional bounces which are only visible in 3D are added,~\eg during motion perpendicular to the camera plane. The bounces separate the trajectory into segments and in each segment we fit a polynomial of degree up to 6. The final trajectory is a function $\mathcal{T}(t)$: $[0,N]\subset\mathbb{R} \to\mathbb{R}^3$ where $N$ is the number of frames. It is defined as
\begin{equation}
	\label{eq:fitting_seq_def}
	\mathcal{T}(t) = \sum_{k=0}^{p_s} \bar{c}_{s,k}t^k ~~~  t \in [t_{s-1}, t_s], s=1..S,
\end{equation}
with $S$ polynomials, where polynomial with index $s$ has degree $p_s$ and it is represented by its coefficient matrix $\bar{c}_s \in \mathbb{R}^{3,p_s+1}$. Time stamps $t_s$ split the whole sequence into intervals between 0 and $N$, such that $0 = t_0 < t_1 < ... < t_{S-1} < t_S = N$. The degree of the polynomial depends on the number of frames it is fitted to; the interpolation scheme is similar to~\cite{tbdnc}.

\subsection{Angular velocity}
\label{sec:rot}
In the case of spherical objects, we are able to estimate their angular velocity $\omega\in\mathbb{R}^3$. Following the standard definition in physics, $\omega$ is a 3D vector of the rotation axis orientation whose magnitude represents the rotation angle along the axis per time unit. Let $\operatorname{\mathbf{R}}_\omega$ be an operator transforming a 2D image of a ball by performing 3D rotation given by $\omega$ of a virtual 3D representation of the ball. More specifically, if $F_2=\operatorname{\mathbf{R}}_\omega F_1$, then $F_2$ is given by mapping the 2D image $F_1$ to a virtual 3D sphere, rotating the sphere by $\omega$ and projecting back on the 2D image. The error of the transformation between the two images is defined as
\begin{equation}
\label{eq:rot_error_def}
E(F_1,F_2|\,\omega) = \|\operatorname{\mathbf{R}}_{\omega}F_1-F_2\|_1.
\end{equation}
Since different parts of the ball are visible before and after rotation, the sum in eq. \eqref{eq:rot_error_def} is carried out only in some pre-defined region visible in both images after arbitrary considered rotation, so that errors corresponding to different rotations are mutually comparable.

Having recovered the object appearance $F_1$ and $F_2$ at two different video sequence timestamps $t_1$ and $t_2$, we can find the average angular velocity $\omega$ between $t_1$ and $t_2$ as the minimizer of the transformation error $E(F_1,F_2|\,(t_2-t_1)\omega)$. Velocity estimation from just two restored images at close timestamps is prone to errors, especially if either of the images is estimated with artifacts. We therefore state an assumption that angular velocity is locally constant in small time interval of the motion (which is physically nearly correct even in the long term if the ball is in free flight) and estimate angular velocity more robustly in a sliding-window manner from several restored images belonging to the corresponding time-window.

Let $F_1,\ldots, F_n$ be a set of estimated ball appearances at timestamps $t_1,\ldots,t_n$; a short time-window of the whole sequence. We estimate a single average angular velocity $\omega$ at this time-window as follows. Let $\omega_{ij}$ be the minimizer of the transformation error from $F_i$ to $F_j$ and $S_{ij}$ inverse of the attained error (`score'):
\begin{align}
\omega_{ij} &= \operatorname*{argmin}_\omega E(F_i,F_j|\,(t_j-t_i)\omega),\label{eq:rot_argmin_omega}\\
S_{ij} &= \frac{1}{E(F_i,F_j|\,(t_j-t_i)\omega_{ij})+\varepsilon}.
\end{align}
In other words, $\omega_{ij}$ is the vote of the corresponding image pair for the true $\omega$ and $S_{ij}$ is the confidence of such vote. We minimize \eqref{eq:rot_argmin_omega} by searching the discretized space of feasible angular velocities.

\input{figures/deformation.tex}

Simply averaging $\omega_{ij}$ results in non-robust estimate that is sensitive to outliers. 
Instead we proceed with RANSAC-like approach. Based on the discretization step used in the minimization of \eqref{eq:rot_argmin_omega}, an inlier threshold $\rho$ is defined as the maximum acceptable error in determining $\omega$. We treat $\omega_{ij}$'s as hypotheses for the final estimate $\omega$ and for each hypothesis calculate its consensus set $C_{ij}$ as
\begin{equation}
C_{ij} = \left\{(k,l)\!: \|\omega_{kl}-\omega_{ij}\| \leq \rho\right\}.
\end{equation}
The winning candidate $\omega_{pq}$ is the one with the best total score of its consensus set,
\begin{equation}
(p,q)=\operatorname*{argmax}_{(i,j)}\sum_{(k,l) \in C_{ij}} S_{kl}.
\end{equation}
The final estimate is then the weighted average of the votes in the consensus set of the winning candidate
\begin{equation}
\omega = \frac{\sum_{kl}S_{kl}\omega_{kl}}{\sum_{kl}S_{kl}},\quad (k,l)\in C_{pq}.
\end{equation}

%% file: tables/comparison.tex
\begin{table*}[t]
\centering
\setlength{\tabcolsep}{1.8mm}
\begin{center}
\begin{tabular}{l|c||c|c|c|c||c|c|c||c|c||c|c}
\hline

\multirow{2}{*}{Sequence} & \multirow{2}{*}{\#}  & \multicolumn{4}{c||}{TIoU-3D}  & \multicolumn{3}{c||}{Radius Error [pixels]} & \multicolumn{2}{c||}{Axis Error [$^{\circ}$]} & \multicolumn{2}{c}{Angle Error [$^{\circ}$]} \\ 
 \cline{3-13}


   &   & {\scriptsize TbD} & {\scriptsize TbD-NC} & {\scriptsize TbD-3D} & {\scriptsize {TbD-3D-O}} & {\scriptsize TbD-NC} & {\scriptsize TbD-3D} & {\scriptsize {TbD-3D-O}} & 
   {\scriptsize TbD-3D} &  {\scriptsize {TbD-3D-O}}  & 
   {\scriptsize TbD-3D}  & {\scriptsize {TbD-3D-O}}  \\ \hline


depthf1 & 46 & .550 & .579 & .625 & \textbf{.937} & 3.348 & 1.333 & \textbf{1.035} & \textbf{59.796} & 60.881 & \textbf{0.124} & 2.269\\  
 depthf2 & 50 & .475 & .528 & .599 & \textbf{.911} & 6.424 & 3.209 & \textbf{1.678} & \textbf{19.966} & 22.125 & 1.733 & \textbf{0.097}\\  
 depthf3 & 37 & .317 & .363 & .452 & \textbf{.763} & 10.986 & 6.004 & \textbf{5.397} & 21.185 & \textbf{11.932} & 1.336 & \textbf{0.556}\\  
 depth2 & 48 & .448 & .590 & .626 & \textbf{.906} & 4.213 & 2.549 & \textbf{1.894} & \textbf{71.085} & 85.816 & \textbf{6.715} & 8.242\\  
 depthb2 & 81 & .366 & .444 & .388 & \textbf{.949} & 2.101 & 7.080 & \textbf{0.850} & \textbf{68.061} & 69.126 & 9.760 & \textbf{7.838}\\  
 out1 & 57 & .465 & .495 & .562 & \textbf{.964} & 6.865 & 2.286 & \textbf{0.705} & 47.329 & \textbf{13.308} & 0.673 & \textbf{0.308}\\  
 out2 & 50 & .503 & .533 & .561 & \textbf{.981} & 4.251 & 1.354 & \textbf{0.369} & \textbf{18.259} & 45.009 & \textbf{0.152} & 0.236\\  
 outb1 & 41 & .350 & .384 & .431 & \textbf{.939} & 4.932 & 3.361 & \textbf{0.885} & 18.856 & \textbf{13.819} & 1.692 & \textbf{0.658}\\  
 outf1 & 60 & .551 & .587 & .611 & \textbf{.968} & 3.297 & 0.924 & \textbf{0.614} & 25.174 & \textbf{12.743} & \textbf{0.015} & 0.041\\  
 \hline 
  Average  & 52 & .447 & .500 & .539 & \textbf{.924} & 5.157 & 3.122 & \textbf{1.492} & 38.857 & \textbf{37.195} & 2.467 & \textbf{2.250}\\

\hline
\end{tabular}
\end{center}
\vspace*{-0.5cm}
\caption{TbD-3D dataset -- comparison of TbD~\cite{tbd}, TbD-NC~\cite{tbdnc} and the proposed TbD-3D. For each sequence, we report: TIoU-3D~\eqref{eq:tiou3d} to measure the accuracy of 3D object location, radius error, axis error as the average angle between the estimated axis and the ground truth axis, and the angle error in degrees. For each sequence and each score, we highlight the best performing method in bold. TbD-3D-O means TbD-3D with oracle: the 2D object location is known from the ground truth. The TbD-3D dataset corresponds to 30 fps frame rate, 8 times lower than the ground truth data from the high speed camera. Results for other frame rates are shown in Figure~\ref{fig:perf_tiou}.}

\label{tbl:comparison}
\end{table*}




%% file: figures/perf_tiou.tex
\begin{figure*}
\centering
\begin{center}
\begin{tabular}{@{}c@{}c@{}c@{}}
\includegraphics[width=0.33\textwidth]{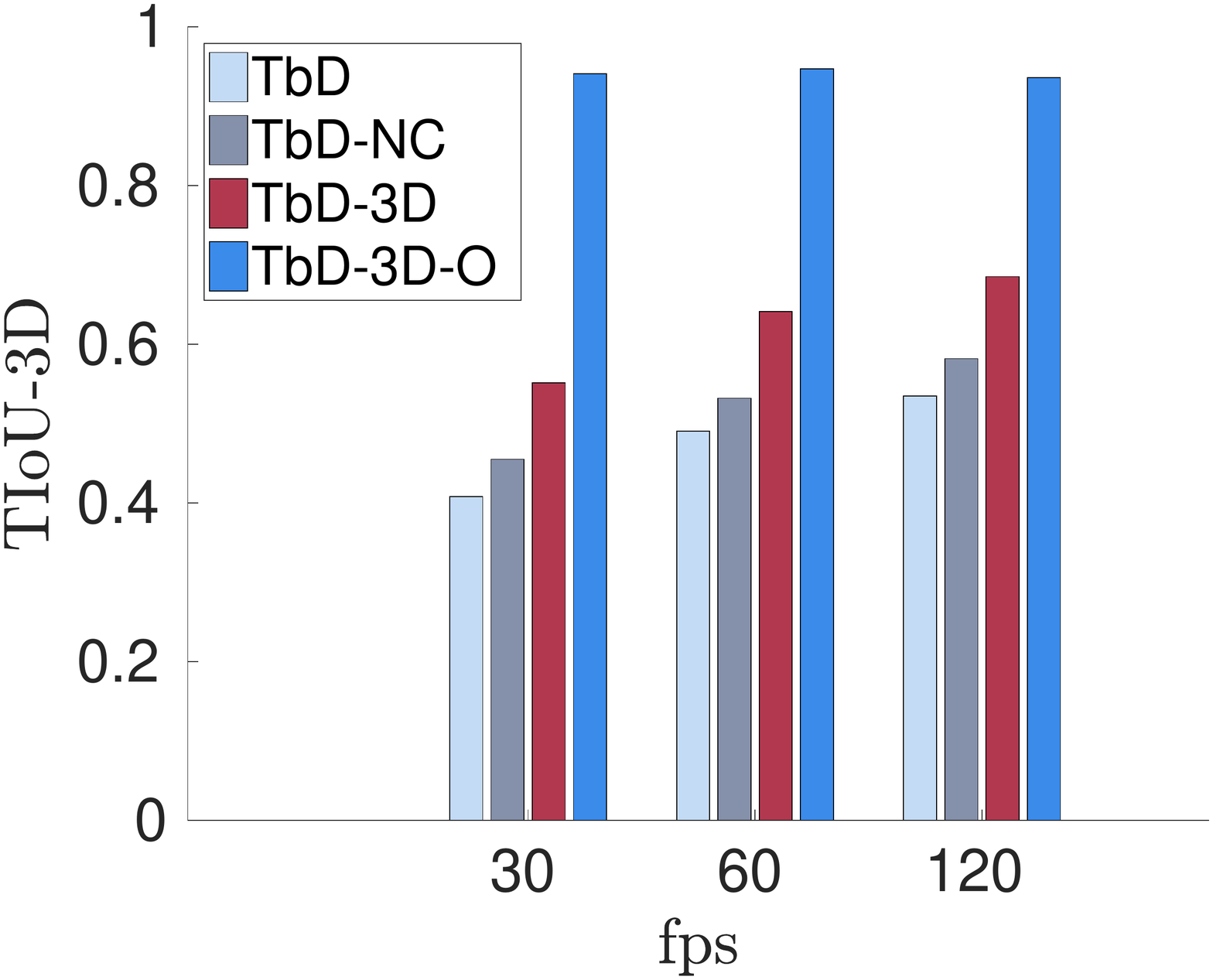} &
\includegraphics[width=0.33\textwidth]{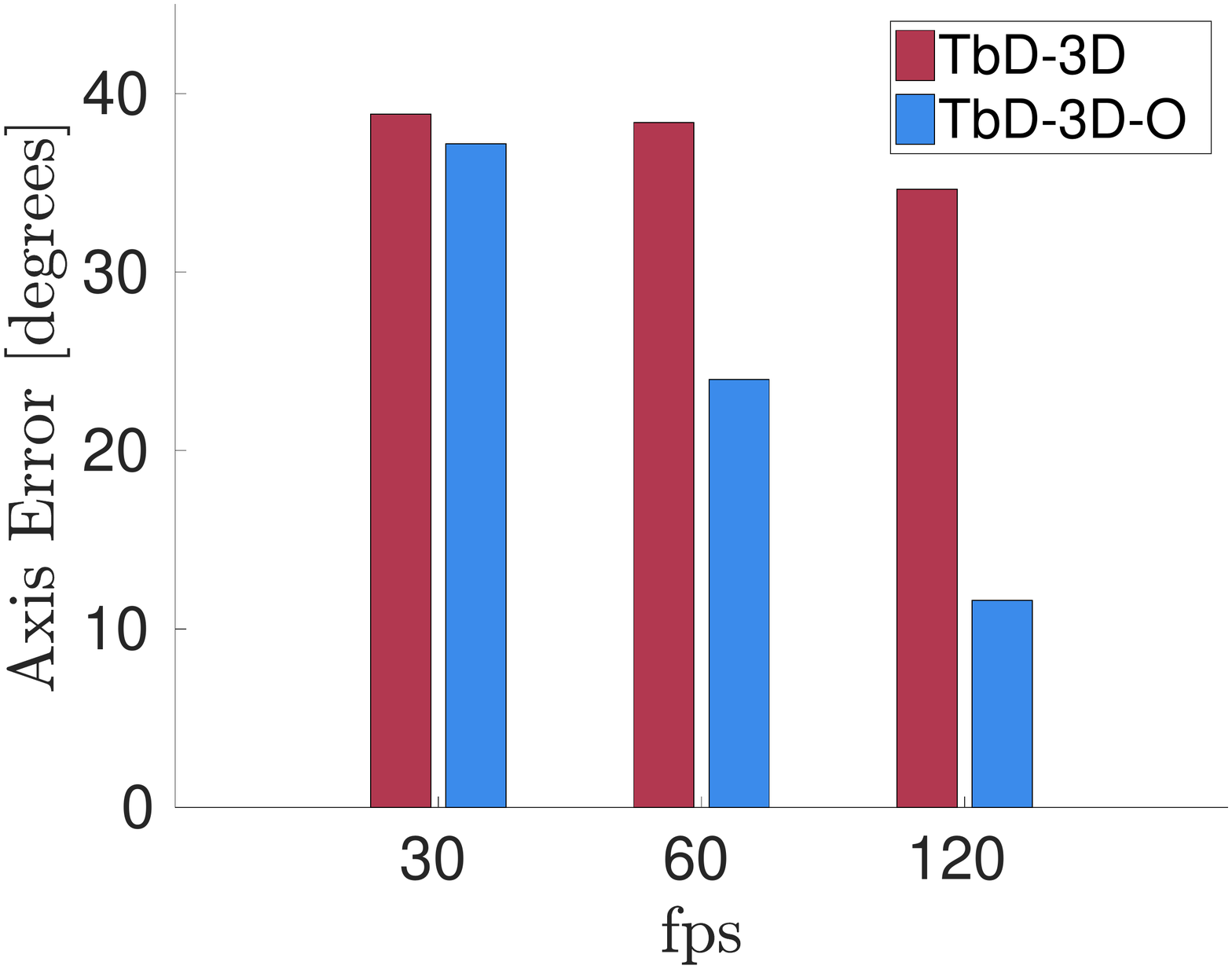} &
\includegraphics[width=0.33\textwidth]{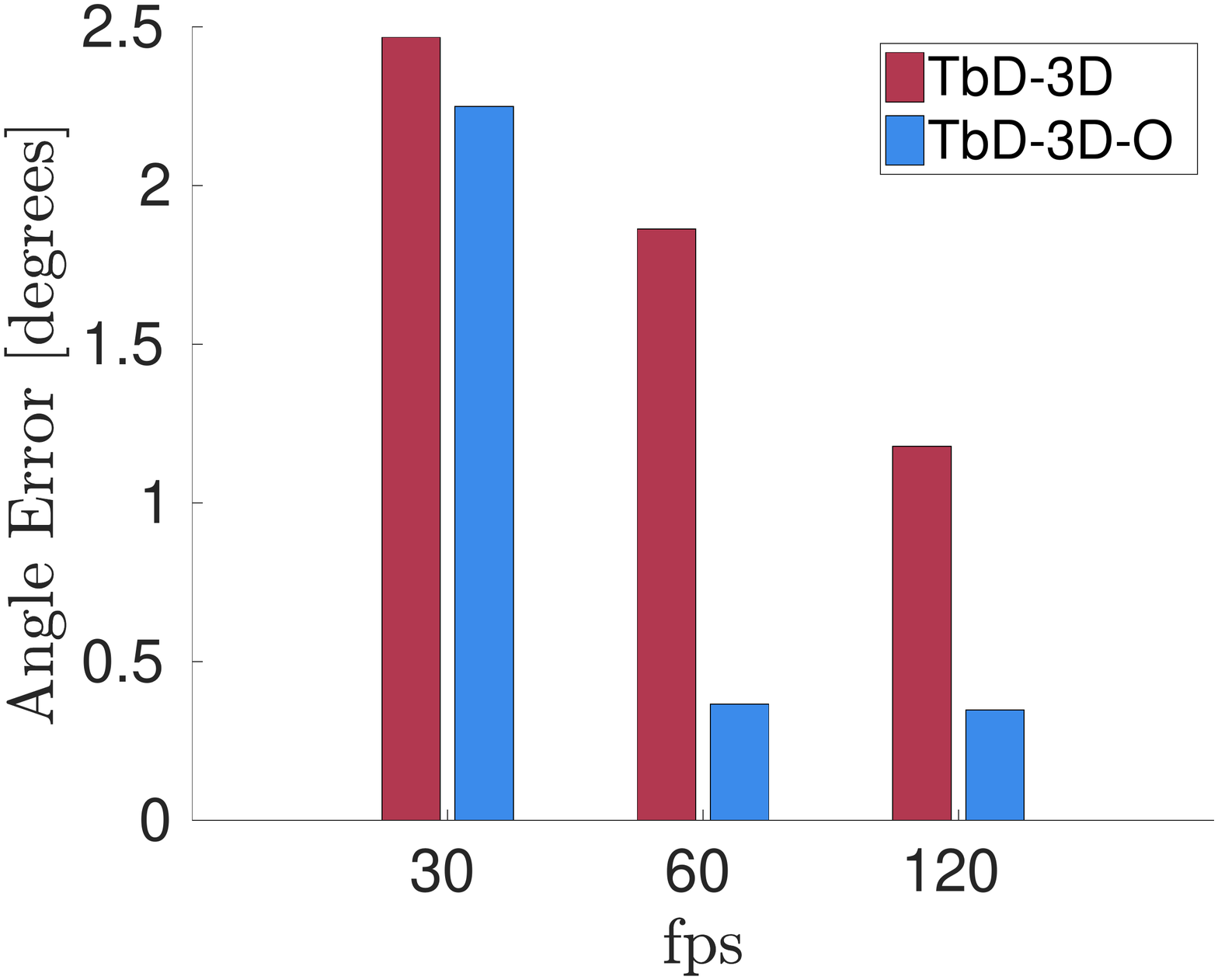} \\


3D Location & 2D Rotation Axis & 1D Rotation Angle \\
\end{tabular}
\end{center}
\vspace*{-0.5cm}
\caption{Evaluation of the TbD-3D method on the TbD-3D dataset with different frame rates. We report scores for three settings: 30, 60 and 120~fps. From left to right: TIoU-3D~\eqref{eq:tiou3d} of the proposed TbD-3D compared to the TbD~\cite{tbd} and TbD-NC~\cite{tbdnc} methods, error of rotation axis estimation, error of rotation angle estimation. The errors of rotation axis and angle are measured by a mean average angle between the estimate and the ground truth from the high-speed footage at 240~fps. Oracle with known 2D trajectory from ground truth is marked by "-O". The TbD-3D method performs better in the task of 3D location estimation and provides meaningful results for 3D rotation estimation w.r.t. the ground truth.
}

\label{fig:perf_tiou}
\end{figure*}

%% file: figures/deformation.tex
\begin{figure*}
\centering
\begin{center}





\renewcommand{\arraystretch}{0.3}
\begin{tabular}{@{}c@{\hspace{0.2em}}c@{}}
\includegraphics[height=0.108\textwidth]{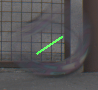} &
\includegraphics[height=0.108\textwidth]{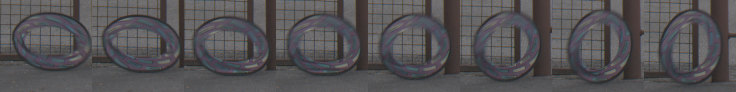} \\
 & \includegraphics[height=0.108\textwidth]{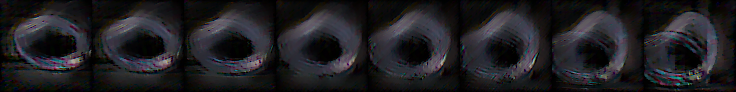} \\
& \includegraphics[height=0.108\textwidth]{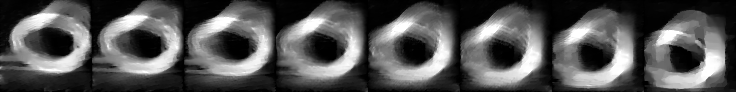} \\[1.0ex]
%
%
%

\multirow{2}{*}[+0.7cm]{\includegraphics[width=0.1\textwidth]{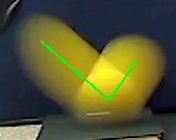}} & 
\includegraphics[width=0.865\textwidth]{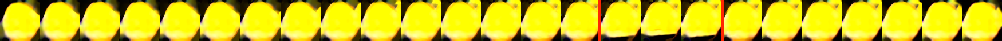} \\
 & \includegraphics[width=0.865\textwidth]{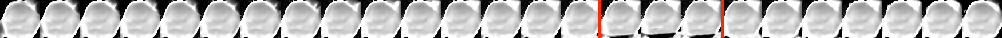} \\

\end{tabular}
\end{center}
\vspace*{-0.4cm}
\caption{Deformations found using the TbD-3D method. They are not modeled explicitly, but are visible during contact with other objects in the scene. Left: input images with trajectories overlaid in green. Right: crops from high-speed camera footage (top), object appearance $F$ and mask $M$ reconstructions by the proposed TbD-3D method with the uniform split of trajectories. For this experiment, we set the term on rotational symmetry $\lambda_R$ in eq.~\eqref{eq:FM_loss_pw} to zero. We estimate sub-frame snapshots using only the input frame on the left and the background. The trajectory is split into 8 (top row) or 25 (bottom row) segments. Deformation during a soft ball bounce is visible between the two red bars in the bottom row.}
\label{fig:deform}
\end{figure*}


%% file: sections/experiments.tex
Kotera~\etal~\cite{tbd} introduced Trajectory Intersection over Union (TIoU) to measure the accuracy of estimated trajectories, which is defined as
\begin{equation}
	\label{eq:tiou}
	\operatorname{TIoU}(\mathcal{C},\mathcal{C}^*) = \int_{t} \operatorname{IoU}\left(\rule{0pt}{2ex}M^*_{\mathcal{C}(t)},\, M^*_{\mathcal{C}^*(t)}\right)\mathrm{d}t,
\end{equation}
where $M^*_{\mathcal{C}(t)}$ corresponds to ground truth object appearance mask $M^*$ placed at a 2D point on either the estimated trajectory $\mathcal{C}(t)$ or ground truth trajectory $\mathcal{C}^*(t)$. Integral is approximated by sum, sampling time at 8 evenly-separated instants. We extend this measure to 3D trajectories and define TIoU-3D as
\begin{equation}
	\label{eq:tiou3d}
	\operatorname{TIoU-3D}(\mathcal{T},\mathcal{T}^*) = \int_{t} \operatorname{IoU}\left(\rule{0pt}{2ex}S^*_{\mathcal{T}(t)},\, S^*_{\mathcal{T}^*(t)}\right)\mathrm{d}t,
\end{equation}
where $S^*_{\mathcal{T}(t)}$ is a ball corresponding to the ground truth radius and located at $\mathcal{T}(t)$, a 3D point along trajectory $\mathcal{T}$ at time-stamp $t$. Similarly, $\mathcal{T}^*$ stands for the ground truth trajectory.

\subsection{TbD-3D Dataset}
We created a new annotated dataset containing fast moving objects. All previous datasets with FMOs, such as FMO dataset~\cite{fmo} and TbD dataset~\cite{tbd}, included only objects moving in a 2D plane parallel to the camera plane and their appearance was close to static. Ground truth 2D object location was provided, but no angular velocity. 

The introduced dataset is the first dataset with non-negligible 3D object motion and with changing appearance of non-uniform fast moving objects. Objects are from a set of three balls with complex texture. The dataset is called TbD-3D and it contains nine sequences with annotated object location, pose, and size from a high-speed camera. In contrast to previous datasets, the perceived size of objects in TbD-3D dataset varies throughout the whole sequence due to depth of the scene, as shown in Figure~\ref{fig:recon2}. 

Videos were recorded in raw format using a high-speed camera at 240~fps with exposure time 1/240s (so called 360$^{\circ}$ shutter angle -- negligible lag between two frames). The dataset sequences were generated by averaging 2, 4 and 8 frames, which corresponds to real videos captured at 30, 60, 120~fps, respectively. Ground truth annotation was done on the original raw footage at 240 fps. 3D object location (2D position and radius) was annotated manually and 3D object rotation was estimated using the proposed method in Section~\ref{sec:rot} and validation; see Section~\ref{sec:exp_rot} for details about ground-truth annotation of the object rotation.

\input{figures/plot_rotation_speed.tex}

The proposed method is evaluated on the TbD-3D dataset for all three frame-rate settings. Figure~\ref{fig:perf_tiou} shows accuracy of the estimated 6DoF object pose: 3D location error measured by TIoU-3D (left), 2D rotation axis error measured as a mean average deviation from the GT axis in degrees (middle) and mean average error of 1D rotation angle (right). We use TbD~\cite{tbd} and TbD-NC~\cite{tbdnc} as baselines, which only estimate 2D trajectory. These methods ignore depth changes and assume one object size for the whole sequence. To show the performance of TbD-3D when the input 2D trajectory has no errors, we also provide scores of TbD-3D with oracle (TbD-3D-O) where we use 2D trajectory from the annotated 240-fps videos. TbD-3D-O estimates only additional 4DoF of object pose and compare to TbD-3D it performs better in average. The performance drop of TbD-3D can be thus attributed to errors in 2D trajectories estimated by TbD-NC. Table~\ref{tbl:comparison} provides more detailed comparison on every sequence at the lowest frame rate of 30~fps. Three examples of 3D trajectory reconstruction on sequences `depth2', `depthf1' and `depthb2' are shown in Figure~\ref{fig:tbd3d_imgs} and one example of angular velocity estimation on sequence `out2' is in Figure~\ref{fig:plot_rotation_speed}. 


\blfootnote{\small \textbf{Acknowledgements.} This work was supported by Czech Science Foundation grant GA18-05360S. Denys Rozumnyi was also supported by Google Focused Research Award.}

\subsection{Rotation Estimation}
\label{sec:exp_rot}
Calculating ground truth rotation of FMOs, even when the high speed camera footage is available, is a challenging task. To estimate the accuracy of the proposed method for rotation estimation (Section~\ref{sec:rot}) when applied on high-speed footage, we captured sequences of a ball rolling on the ground along a straight trajectory of known length. The ground truth angular velocity is derived from physical properties of the rolling ball as we know its actual circumference and the distance it traverses. The average angle between the estimated rotation axis using the proposed method and the GT axis was 4.052 degrees. The average angle between the estimated and GT rotation angle was 0.037 degrees, which corresponds to 1.2 \% relative error.

A special case appears during contact with another object in the scene. The object is deformed and modeling the object there is out of the scope of this paper. However, we can still detect such deformations as shown in Figure~\ref{fig:deform}.

\subsection{Applications}
Temporal super-resolution is among the most interesting applications of the proposed method. First, a video free of FMOs is produced by replacing blurred objects with the estimated background. Second, a higher frame rate video is created by linear interpolation. Last, the trajectory is split into the desired number of segments and the object is blended into the sequence with its 6DoF appearance at desired snapshot time-scale, following the image formation model~\eqref{eq:acquisition_model}. Compared to the previous methods, which use the same appearance for all frames among one low rate trajectory, we synthesize the object at much higher temporal resolution. Videos generated using temporal super-resolution are provided in the supplementary material.

Other applications and future work include for instance rotation estimation for tennis or table tennis serves, or full 3D reconstruction of the blurred object.

%% file: figures/plot_rotation_speed.tex
\begin{figure}
\centering
\includegraphics[width=\columnwidth]{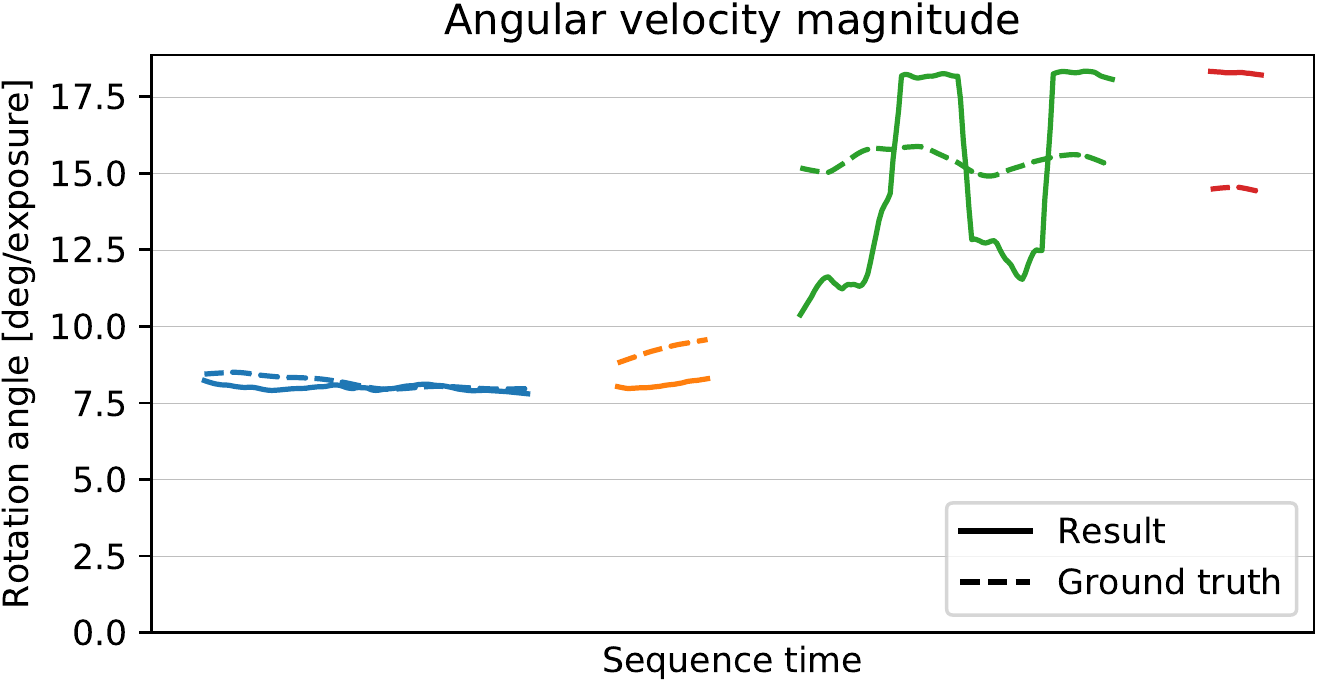}
\includegraphics[width=\columnwidth]{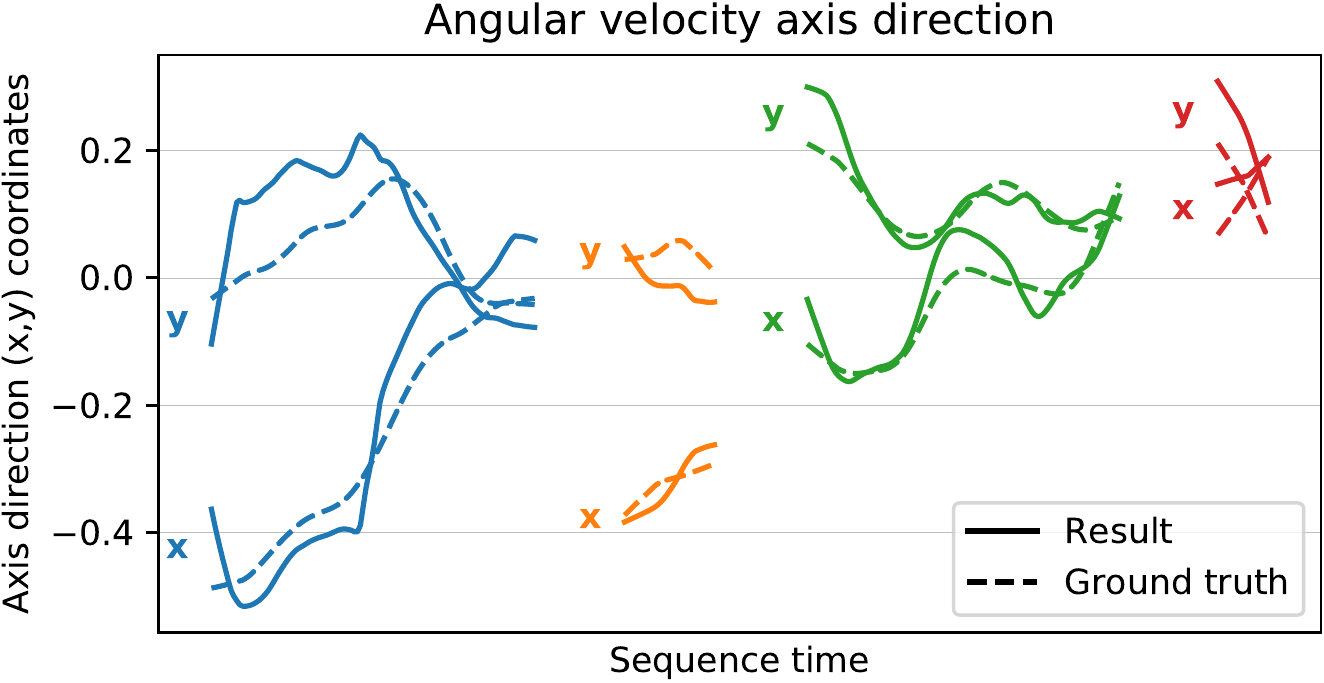}
\includegraphics[width=\columnwidth]{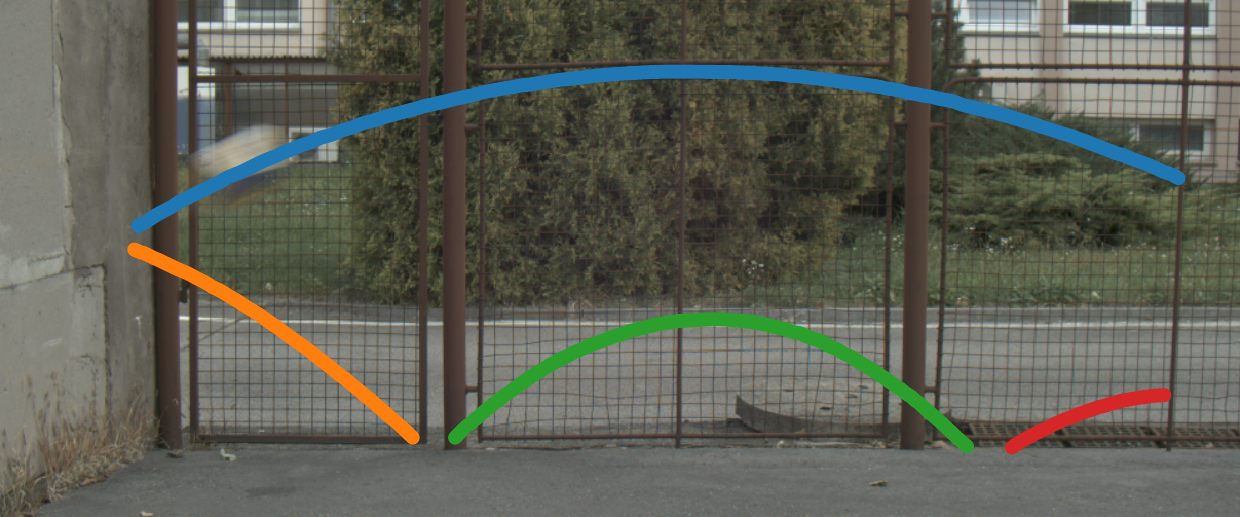}
\vspace*{-0.5cm}
\caption{Rotation velocity magnitude and direction in different parts of the sequence (color coded). TbD-3D results~--~solid lines, ground truth~--~dashed. Rotation is estimated only between bounces.}
\label{fig:plot_rotation_speed} 
\end{figure}